\documentclass{article} 
\usepackage{iclr2024_conference,times}

\usepackage{amsmath,amsfonts,bm}

\def\eqref#1{equation~\ref{#1}}

\def\1{\bm{1}}

\DeclareMathAlphabet{\mathsfit}{\encodingdefault}{\sfdefault}{m}{sl}
\SetMathAlphabet{\mathsfit}{bold}{\encodingdefault}{\sfdefault}{bx}{n}

\usepackage{xcolor}
\usepackage{hyperref}
\hypersetup{
    colorlinks=true,
    linkcolor=blue!75!black,
    citecolor=blue!75!black
}
\usepackage{url}
\usepackage[utf8]{inputenc}
\usepackage[htt]{hyphenat}
\usepackage{authblk}
\usepackage{algorithm}
\usepackage{algpseudocode}
\usepackage{longtable}
\usepackage{rotating}
\usepackage{listings}
\usepackage{tabularx}
\usepackage{amsmath}
\usepackage{xcolor}
\usepackage{booktabs}
\usepackage{graphicx}
\usepackage{tikz}
\usetikzlibrary{positioning}
\usepackage{svg}
\usepackage{cleveref}
\usepackage{courier}
\usepackage{seqsplit}
\usepackage[inline]{enumitem}
\usepackage{multirow}
\usepackage[most]{tcolorbox}
\usepackage{float}

\DeclareMathOperator{\LLM}{LLM}

\newcommand{\cen}[1]{\multicolumn{1}{c}{\textbf{#1}}}
\newcommand{\h}[1]{\textbf{#1}}

\newcommand{\prompt}[1]{\footnotesize{\texttt{"#1"}}\normalsize}

\title{Promptbreeder:\\Self-Referential Self-Improvement\\ via Prompt Evolution}

\author{Chrisantha Fernando, Dylan Banarse, Henryk Michalewski, Simon Osindero, Tim Rocktäschel\\
Google DeepMind\\
\texttt{\{chrisantha,dylski,henrykm,osindero,rocktaschel\}@google.com}
}

\iclrfinalcopy 

\renewcommand{\headrulewidth}{0.5pt}
\fancypagestyle{firstpage}{
  \lhead{\begin{picture}(0,0)\put(0,-3){\includegraphics[width=0.25\linewidth]{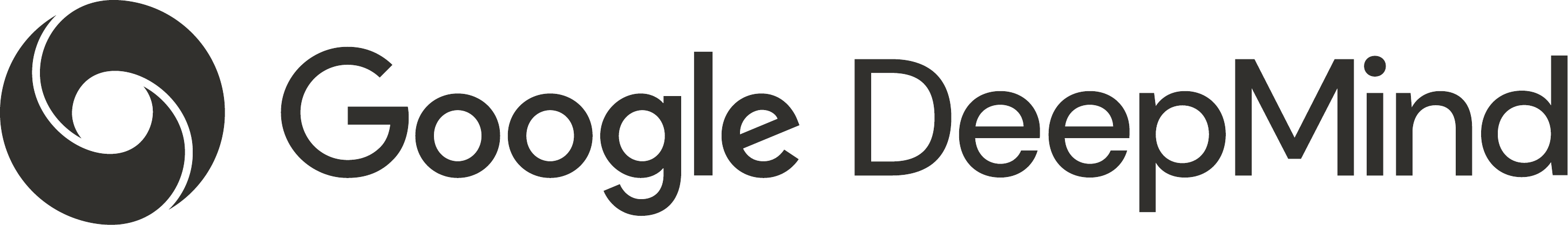}}\end{picture}}
}

\begin{document}

\maketitle
\thispagestyle{firstpage}
\vspace{-1em}

\begin{abstract}
Popular prompt strategies like Chain-of-Thought Prompting can dramatically improve the reasoning abilities of Large Language Models (LLMs) in various domains. However, such hand-crafted prompt-strategies are often sub-optimal. In this paper, we present \textsc{Promptbreeder}, a general-purpose self-referential self-improvement mechanism that evolves and adapts prompts for a given domain. Driven by an LLM, Promptbreeder mutates a population of task-prompts, evaluates them for fitness on a training set, and repeats this process over multiple generations to evolve task-prompts. Crucially, the mutation of these task-prompts is governed by mutation-prompts that the LLM generates and improves throughout evolution in a self-referential way. That is, Promptbreeder is not just improving task-prompts, but it is also improving the mutation-prompts that improve these task-prompts. Promptbreeder outperforms state-of-the-art prompt strategies such as Chain-of-Thought and Plan-and-Solve Prompting on commonly used arithmetic and commonsense reasoning benchmarks. Furthermore, Promptbreeder is able to evolve intricate task-prompts for the challenging problem of hate speech classification.
\end{abstract}

\section{Introduction}

Prompting is central to the downstream performance of foundation models. For example, different prompt strategies\footnote{See \Cref{appendix:glossary} for definitions of terminology.} can have a significant impact on a model's reasoning abilities~\citep{DBLP:conf/nips/Wei0SBIXCLZ22,DBLP:journals/corr/abs-2112-00114,zhou2022least,wang2022self,zhou2022large,DBLP:conf/acl/WangXLHLLL23}, multi-modal processing abilities~\citep{yang2023mm,DBLP:journals/corr/abs-2302-01560}, or tool use abilities~\citep{yao2022react,schick2023toolformer}. Furthermore, prompting can improve model distillation~\citep{DBLP:conf/acl/WangKMLSKH23,DBLP:conf/acl/HsiehLYNFRKLP23} and it can be used to simulate agentic behavior~\citep{DBLP:journals/corr/abs-2305-16291,DBLP:journals/corr/abs-2304-03442,DBLP:journals/corr/abs-2305-15486}. However, these prompt strategies are manually engineered. Since the specific way a prompt is phrased can have a dramatic effect on its utility~\citep{DBLP:journals/corr/abs-2209-07686}, it raises the question of whether prompt engineering can be automated. Automatic Prompt Engineer \citep[APE,][]{zhou2022large} attempts to address this by generating an initial distribution of prompts using another prompt that infers the problem from a number of input-output examples from the dataset. However, \citeauthor{zhou2022large} found ``diminishing returns to further selection rounds as the quality seems to stabilize after three rounds'', and consequently abandoned the use of an iterative APE. We propose a solution to the problem of diminishing returns via a diversity maintaining evolutionary algorithm for self-referential self-improvement of prompts for LLMs.  

\citet{schmidhuber1990making} notes that the ``program of a neural network is its weight matrix''. Consequently, this ``program'' can be changed in a self-referential way by the neural network itself~\citep{schmidhuber1993selfreferential,DBLP:conf/icml/IrieSCS22}. Such a neural network that improves itself, as well as improving the way it improves itself, might be an important stepping stone towards open-ended self-referential self-improvement of AIs~\citep{schmidhuber2003godel}. However, self-improvement via self-referential weight matrices is costly as it requires additional parameters that modify all of the model's parameters. Since behaviors and capabilities of LLMs are significantly influenced by the prompts that we provide to them, we can similarly think of prompts as the program of an LLM~\citep{zhou2022large}. In this view, changing a prompt strategy such as the Scratchpad method~\citep{DBLP:journals/corr/abs-2112-00114} or Chain-of-Thought Prompting~\citep{DBLP:conf/nips/Wei0SBIXCLZ22} corresponds to changing the ``program'' of the LLM. Taking this analogy further, we can use the LLM itself to change its prompts, as well as the way it changes these prompts, moving us towards a fully self-referential self-improving systems grounded in LLMs.

\newcommand{\row}[8]{ #1 & #5 & #3 & #6 & #8 & #7 & #4 & #2\\}
\begin{table}[t!]
\centering
\resizebox{\textwidth}{!}{
\begin{tabular}{rrr|rrrr|rrrr}
\toprule
 & \cen{Method} & \cen{LLM} & \row{\cen{MultiArith*}}{\cen{GSM8K}}{\cen{AddSub*}}{\cen{AQuA-RAT}}{\cen{SingleEq*}}{\cen{SVAMP*}}{\cen{CSQA}}{\cen{SQA}}
\midrule
\parbox[t]{2mm}{\multirow{9}{*}{\rotatebox[origin=c]{90}{Zero-shot}}} & CoT & text-davinci-003 & \row{(83.8)}{(56.4)}{(85.3)}{(38.9)}{(88.1)}{(69.9)}{(65.2)}{(63.8)}
 & PoT & text-davinci-003 & \row{(92.2)}{(57.0)}{(85.1)}{(43.9)}{(91.7)}{(70.8)}{--}{--}
 & PS & text-davinci-003 & \row{(87.2)}{(58.2)}{(88.1)}{(42.5)}{(89.2)}{(72.0)}{--}{--}
 & PS+ & text-davinci-003 & \row{(91.8)}{(59.3)}{(\h{92.2})}{(46.0)}{(94.7)}{(75.7)}{(71.9)}{(65.4)}
& PS & PaLM 2-L & \row{97.7}{59.0}{72.4}{40.2}{90.6}{83.8}{77.9}{50.0}
& PS+ & PaLM 2-L & \row{92.5}{60.5}{74.4}{39.4}{94.7}{86.3}{73.3}{50.1}
& APE & PaLM 2-L &  \row{95.8}{77.9}{72.2}{45.7}{82.2}{73.0}{67.3}{38.4}
& OPRO & PaLM 2-L &  \row{--}{80.2}{--}{--}{--}{--}{--}{--}
 & PB (ours)  & PaLM 2-L & \row{\h{99.7}}{\h{83.9}}{87.8}{\h{62.2}}{\h{96.4}}{\h{90.2}}{\h{85.4}}{\h{71.8}} 
\midrule
\parbox[t]{2mm}{\multirow{3}{*}{\rotatebox[origin=c]{90}{Few-}}} & Manual-CoT & text-davinci-003 & \row{(93.6)}{(58.4)}{(\h{91.6})}{(48.4)}{(93.5)}{(80.3)}{(78.3)}{(71.2)}
 & Auto-CoT & text-davinci-003 & \row{(95.5)}{(57.1)}{(90.8)}{(41.7)}{(92.1)}{(78.1)}{--}{--}        
 & PB (ours) & PaLM 2-L & \row{\h{100.0}}{\h{83.5}}{87.1}{\h{64.6}}{\h{98.9}}{\h{93.7}}{\h{85.9}}{\h{80.2}} 
\bottomrule
\end{tabular}
}
\caption{Promptbreeder (\textbf{PB}) comparison to Chain-of-Thought~\citep[\textbf{Manual-CoT},][]{DBLP:conf/nips/Wei0SBIXCLZ22}, Zero-shot \textbf{CoT}~\citep{DBLP:conf/nips/KojimaGRMI22}, Program-of-Thoughts~\citep[\textbf{PoT,}][]{chen2022program}, \textbf{Auto-CoT}~\citep{DBLP:conf/iclr/0001Z0S23}, \textbf{OPRO}~\citep{DBLP:journals/corr/abs-2309-03409}, Automatic Prompt Engineer Zero-shot prompt~\citep[\textbf{APE},][]{zhou2022large}, Plan-and-Solve with (\textbf{PS+}) and without the improved prompt~\citep[\textbf{PS},][]{DBLP:conf/acl/WangXLHLLL23} and using PaLM 2-L~\citep{anil2023palm} as the underlying LLM (\textbf{APE}, \textbf{PS}$_\textbf{PaLM 2-L}$/\textbf{PS}+$_\textbf{PaLM 2-L}$). Best results in both the zero-shot and few-shot categories are highlighted in bold. Results in brackets are directly taken from the Plan-and-Solve paper which uses text-davinci-003~ \citep{DBLP:conf/nips/BrownMRSKDNSSAA20}. For datasets with astericks (MultiArith*, SingleEq*, AddSub*, and SVAMP*), we randomly took half of the examples for training and report accuracy on the remaining test set. See \Cref{sec:experiments} and \Cref{appendix:datasets} for details on the prompts and datasets.}
\label{tab:pe_comparison}
\end{table}

In this paper, we introduce \textsc{Promptbreeder} (PB) for self-referential self-improvement of LLMs. Given a seed set of mutation-prompts (i.e. instructions to modify a task-prompt), thinking-styles (i.e. text descriptions of general cognitive heuristics), and a domain-specific problem description, PB generates variations of the task-prompts and mutation-prompts, exploiting the fact that LLMs can be prompted to act as mutation operators~\citep{DBLP:journals/corr/abs-2302-12170}. Based on the fitness of the evolved task-prompts as measured on the training set, we select a subset of evolutionary units consisting of task-prompts and their associated mutation-prompt, to transmit to future generations. Over multiple generations of PB, we observe prompts adapting to the domain at hand. For example, in a mathematical domain, PB evolved the task-prompt \prompt{Show all your working. II. You should use the correct mathematical notation and vocabulary, where appropriate. III. You should write your answer in full sentences and in words. IV. You should use examples to illustrate your points and prove your answers. V. Your workings out should be neat and legible} on GSM8K (see \Cref{appendix:examples}). On a wide range of commonly used benchmarks spanning commonsense reasoning, arithmetic, and ethics, we find that PB outperforms state-of-the-art methods like Chain-of-Thought~\citep{DBLP:conf/nips/Wei0SBIXCLZ22} and Plan-and-Solve~\citep{DBLP:conf/acl/WangXLHLLL23} prompting. 
As PB does not require any parameter updates for self-referential self-improvement, we believe this approach points to an interesting future where larger and more capable LLMs could further amplify the gains of our approach.

In summary, this paper makes the following main contributions: 
(i) we introduce Promptbreeder, a self-referential self-improvement method for LLMs that evolves prompts for a domain at hand, as well as improves the way it is evolving these prompts,
(ii) we report improvements over state-of-the-art prompt strategies on a wide range of commonly used arithemic and commonsense reasoning benchmarks, and 
(iii) we investigate the various self-referential components of Promptbreeder and their contribution to our results. 

\begin{figure}[t!]
    \centering
    \includegraphics[width=\textwidth]{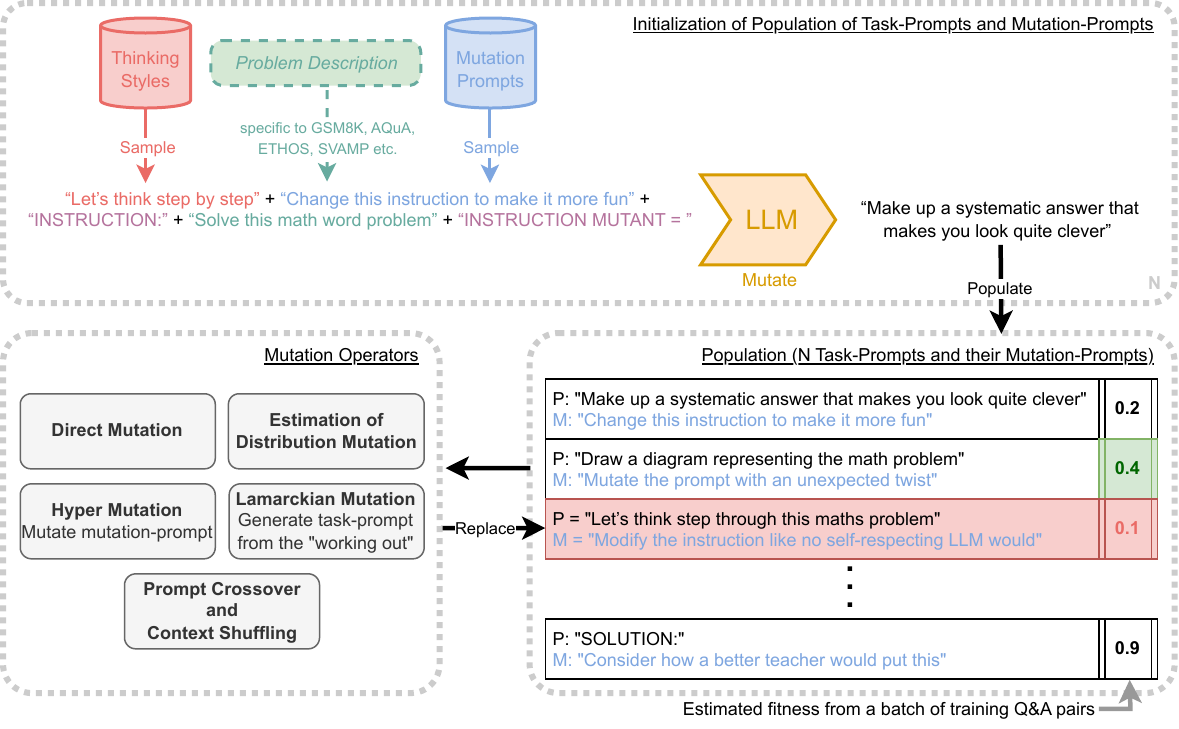}
    \caption{Overview of Promptbreeder. Given a problem description and an initial set of general ``thinking-styles'' and mutation-prompts, Promptbreeder generates a population of units of evolution, each unit consisting of typically two task-prompts and a mutation-prompt. We then run a standard binary tournament genetic algorithm~\citep{harvey2011microbial}. To determine the fitness of a task-prompt we evaluate its performance on a random batch of training data. Over multiple generations, Promptbreeder subsequently mutates task-prompts as well as mutation-prompts using five different classes of mutation operators. The former leads to increasingly domain-adaptive task-prompts whereas the latter evolves increasingly useful mutation-prompts in a self-referential way.}
    \label{fig:overview}
\end{figure}

\section{Related Work}

Prompting an LLM in the right way is essential to its downstream performance~\citep{DBLP:conf/emnlp/MoradiS21,DBLP:journals/corr/abs-2209-07686,zhou2022large}. Indeed, even the order in which prompts are presented can heavily influence LLM performance~\citep{DBLP:conf/acl/LuBM0S22}. A number of recent works have focused on devising better prompt strategies, or even automating such prompt engineering.

\textbf{Prompting}:
Chain-of-Thought Prompting~\citep[CoT,][]{DBLP:conf/nips/Wei0SBIXCLZ22} is a popular prompt strategy which provides intermediate reasoning steps as few-shot prompts to an LLM, thereby significantly improving its arithmetic, commonsense, and symbolic reasoning abilities. Notably, the gains of CoT are more pronounced for stronger LLMs. This is intriguing, as it points to the possibility of increasingly capable (and potentially open-ended) self-improving mechanisms on top of adept LLMs---a hypothesis that Promptbreeder directly builds upon. Instead of few-shot CoT prompting, \cite{DBLP:conf/nips/KojimaGRMI22} demonstrate that LLMs can also be prompted zero-shot (e.g. \prompt{Let's think step by step}) to produce their own chains of thoughts (Zero-shot CoT) that improve reasoning abilities. Self-Consistency~\citep[CoT-SC,][]{wang2022self} extends CoT by sampling a diverse set of workings out and selecting the most consistent answer. Tree of Thoughts~\citep[ToT,][]{yao2023tree} generalizes CoT to multiple workings out that can be expanded or backtracked from. Graph of Thoughts~\citep[GoT,][]{DBLP:journals/corr/abs-2308-09687} is a further generalization to arbitrary graph structures. Plan-and-Solve Prompting~\citep[PS,][]{DBLP:conf/acl/WangXLHLLL23} encourages an LLM to first devise a plan to solve a problem before attempting to solve it. Similarly, Least-to-Most Prompting~\citep{zhou2022least} encourages an LLM to decompose a problem into subparts, and then to solve each part individually before synthesizing an answer. Self-Refine~\citep{DBLP:journals/corr/abs-2303-17651} prompts an LLM to generate a response, to provide feedback on the response, and to finally refine the solution.

In contrast to gradient-free approaches above, Soft Prompting approaches~\citep[e.g.,][]{DBLP:journals/corr/abs-2103-10385,qin2021learning,DBLP:conf/emnlp/LesterAC21} directly fine-tune continuous prompt representations. \citet{DBLP:journals/corr/abs-2210-11610} use CoT and CoT-SC on an unlabelled dataset of questions, and subsequently fine-tune an LLM based on generated solutions. Similarly, \citet{DBLP:conf/nips/ZelikmanWMG22} uses CoT to generate rationales and fine-tunes the LLM based on those examples and rationales that yielded the correct answer. However, as argued by~\citet{zhou2022large}, any approach that updates all or a portion of LLM parameters will not scale as models get bigger and, moreover, will not work with the increasing number of LLMs hidden behind an API.

All of the prompt engineering approaches above are domain agnostic but hand designed. Central to our work is the hypothesis that we could do better by employing an automated self-improvement process that can adapt prompts to a domain at hand. Auto-CoT~\citep{DBLP:conf/iclr/0001Z0S23} and Automatic-CoT~\citep{DBLP:journals/corr/abs-2302-12822} automatically find reasoning chains for Few-Shot CoT. Automatic Prompt Engineer~\citep[APE,][]{zhou2022large} uses one generator-prompt to generate prompt candidates, and another mutation-prompt to mutate them. In contrast to APE, our work performs compositional task-specific initialization of mutation-prompts, subsequent online mutation of mutation-prompts, uses special mutation operators that take into account the whole population and elite history, and uses diversity-maintenance methods---all of which help avoid the problem of diminishing returns and diversity loss suffered by APE. 

Concurrently to our work, \cite{DBLP:journals/corr/abs-2309-03409} developed Optimization by PROmpting (OPRO), a prompt optimization method that varies prompts using a single complex mutation prompt, and evaluates newly generated prompts on a small fixed training set of problems. In contrast, Promptbreeder autonomously evolves multiple LLM generated mutation-prompts as well as task-prompts, and evaluates fitness on random subsets from the whole training set during evolution. At the time of its release, OPRO achieved a score of 80.2\% via the optimized zero-shot prompt \prompt{Take a deep breath and work on this problem step-by-step} on GSM8K. Promptbreeder surpasses this with 83.9\% in the zero-shot setting with the unintuitively simple prompt \prompt{SOLUTION"}---further evidence for the sensitivity of LLMs to prompts and the importance on finding effective prompts automatically. Also concurrently to our work, \cite{guo2023connecting} developed EvoPrompt, which uses a fixed mutation (and crossover) prompt, as well as a prompt that asks for a mutant of the difference between two parent prompts, to produce offspring prompts. EvoPrompt is initialized with a whole population of initial hand-designed task tailored prompts rather than a single problem description as we do. In contrast to the two approaches above, Promptbreeder uses LLMs to self-referentially improve mutation-prompts, and it is able to evolve contexts as well. 

\textbf{Self-Referential Self-Improvement}: 
Developing an open-ended system that can improve itself as well as improving the way it is improving itself~\citep{schmidhuber1993selfreferential,schmidhuber2003godel} is a long-standing open problem in AI research. \citet{schmidhuber1993selfreferential} introduced an ``introspective'' neural network with a self-referential weight matrix that can modify its own weights and, thus, also modify those weights that are governing how its own weights are modified. Recently, \citet{DBLP:conf/icml/IrieSCS22} proposed a more scalable self-referential weight matrix taking inspiration from fast weight programmers~\citep{schmidhuber1992learning}. \citet{DBLP:journals/corr/abs-2212-14392} propose a self-referential meta-learning approach, combining self-referential weight matrices with ideas from Gödel Machines~\citep{schmidhuber2003godel}, i.e., to allocate more computational resources to better performing solutions. However, since these approaches directly modify parameters of a model, it is unclear how to scale them to the increasing number of parameters in modern LLMs. In contrast, for Promptbreeder the substrate of self-referential self-improvement is natural language, avoiding costly parameter updates altogether.

\textbf{Open-Endedness and LLMs}: 
Promptbreeder makes use of the observation by \cite{DBLP:journals/corr/abs-2206-08896}, \cite{DBLP:journals/corr/abs-2302-12170} and \cite{DBLP:journals/corr/abs-2302-14838} that LLMs are effective at generating mutations from examples. In addition, LLMs encode human notions of interestingness and can be used to automatically quantify novelty \citep{DBLP:journals/corr/abs-2306-01711}. Promptbreeder is related to Picbreeder~\citep{secretan2008picbreeder}, an open-ended human-in-the-loop system that evolves increasingly interesting images. While Picbreeder explores the space of images, Promptbreeder explores the space of prompts and does so without humans in the loop. As Promptbreeder is proposing mutated prompts to itself, it is an example of a system transitioning from ``learning from data'' to ``learning what data to learn from''~\citep{DBLP:journals/corr/abs-2211-07819}.

\section{Promptbreeder}
We introduce Promptbreeder, a prompt evolution system that can automatically explore prompts for a given domain and that is able to find task-prompts that improve an LLM's ability to derive answers to questions in that domain. Promptbreeder is general purpose in that the same system is able to adapt to many different domains. 

Promptbreeder makes use of the observation that LLMs can be used to generate variations of input text~\citep{DBLP:journals/corr/abs-2206-08896,DBLP:journals/corr/abs-2302-12170,DBLP:journals/corr/abs-2302-14838}.  \Cref{fig:overview} gives an overview of our method. We are interested in evolving task-prompts. A task-prompt $P$ is a string used to condition the context of an LLM in advance of some further input $Q$, intended to ensure a better response than if $Q$ had been presented in the absence of $P$. To evaluate the fitness of each evolved task-prompt, we sample a batch of 100 Q\&A pairs from the entire training set of the domain at hand.\footnote{Our prompt strategy sequentially applies two task-prompts. The first task-prompt + question produces a continuation. The continuation + second task-prompt produces the final answer.} 

Promptbreeder generates task-prompts according to an evolutionary algorithm. The mutation operator for this algorithm is itself an LLM, conditioned on a mutation-prompt $M$. That is, a mutated task prompt $P'$ is defined by $P' = \LLM(M + P)$ where `$+$` corresponds to string concatenation. A variety of such mutation-prompts are described in \Cref{sec:mutation}. 

Promptbreeder's main self-referential mechanism stems from applying the evolutionary algorithm not just to task-prompts but also to mutation-prompts. The mutation operator for this meta-level algorithm is again an LLM, now conditioned on a hyper-mutation prompt $H$. That is, we obtain a mutated mutation-prompt $M'$ via $M' = \LLM(H + M)$. 

Given a set of ``thinking styles" $\mathcal{T}$ and a set of initial mutation-prompts $\mathcal{M}$, as well as a domain-specific problem description $D$, Promptbreeder initializes a population of mutated task-prompts (see \Cref{sec:initialization}). To clarify, a unit of evolution consists of a set of task-prompts, a mutation-prompt and in the few-shot case, a set of correct workings out (i.e. step-by-step or ``chains-of-thought'' reasoning steps that led to the correct answer). This means task-prompts and mutation-prompts are in 1:1 correspondence. To evolve this population, we employ a binary tournament genetic algorithm framework \citep{harvey2011microbial}: we sample two individuals from the population, we take the individual with the higher fitness, mutate it (see next section) and overwrite the loser with the mutated copy of the winner.
 
\subsection{Promptbreeder Initialization}
\label{sec:initialization}
To give a concrete example, consider the initialization steps used to produce the task-prompts and mutation-prompts for GSM8K (a `grade school maths' word problem dataset). The problem description is \prompt{Solve the math word problem, giving your answer as an arabic numeral}. Because Plan-and-Solve~\citep{DBLP:conf/acl/WangXLHLLL23} uses two task-prompts we also evolve two task-prompts (plus a mutation-prompt) per unit of evolution. In order to promote diversity in the initial prompts, we generate the initial task-prompts by concatenating (for each task-prompt) a randomly drawn `mutation-prompt' (e.g. \prompt{Make a variant of the prompt.}) and a randomly drawn `thinking-style'  (e.g. \prompt{Let's think step by step}) to the problem description, and provide that to the LLM to produce a continuation, resulting in an initial task-prompt. We do this twice to produce the two initial task-prompts per unit. Both the mutation-prompt and the thinking-style are randomly sampled from an initial set of mutation-prompts and a set of thinking-styles (see Appendices \ref{appendix:mutation}, \ref{appendix:thinking} and \ref{appendix:problem_specifications} for the full sets). The mutation-prompt is added to the unit of evolution and so is associated with its specific task-prompt throughout the evolutionary run. 

For the example above, the complete input string to the LLM to make an initial task-prompt could be \prompt{Make a variant of the prompt. Let's think step by step. INSTRUCTION: Solve the math word problem, giving your answer as an arabic numeral. INSTRUCTION MUTANT:}. Note how the control strings \prompt{INSTRUCTION} and \prompt{INSTRUCTION MUTANT} are added to encourage an appropriate continuation. \Cref{tab:initialization} in \Cref{appendix:initialization} shows examples of the initial prompts generated in this way. 

\subsection{Mutation Operators}
\label{sec:mutation}
As shown in~\Cref{fig:overview}, there are nine operators falling into five broad classes which drive the exploration of prompt strategies. For each replication event only one of nine mutation operators is applied (we sample with uniform probability over the nine operators to decide which mutation operator to apply). The rationale for using this diverse set of operators is to enable the LLM to explore a large space of cognitive methods of linguistic self-questioning, by repeatedly changing the framing of the problem as well as retrieving mental models expressed in natural language that can help tackle a given reasoning challenge. Investigations from insight learning strongly suggest that diverse representational re-description is key to problem solving \citep{ollinger2009psychological}---a principle that we attempt to recreate via self-referential self-improvement with natural language as the substrate. \Cref{fig:selfref} illustrates in what way Promptbreeder is self-referential (see \Cref{appendix:selfref} for a more detailed explanation).

\subsubsection{Direct Mutation}
\label{sec:direct}
The simplest class of mutation operators directly generate a new task-prompt $P'$ from either one existing task-prompt $P$ (first-order prompt generation) or from a general prompt that encourages free-form generation of new task-prompts--i.e. not using an existing parent, thus zero-order prompt generation.

\textbf{Zero-order Prompt Generation}: We generate a new task-prompt by concatenating the problem description $D$ (e.g. \prompt{Solve the math word problem, giving your answer as an arabic numeral}) with the prompt \prompt{A list of 100 hints:}, which invites the LLM to come up with a new hint that could help solve a problem in the given problem domain. We extract the first generated hint as the new task-prompt. Crucially, this new task-prompt does not depend on any previously found task-prompt. Instead, it is re-generated from the problem description each time. Our rationale for including this zero-order operator is that where prompt evolution diverges, this operator allows us to generate new task-prompts closely related to the original problem description, similar to uniform re-sampling in automated curriculum learning approaches~\citep{DBLP:conf/icml/JiangGR21,DBLP:conf/nips/JiangDPFGR21,DBLP:journals/corr/abs-2304-03442, DBLP:conf/icml/Parker-HolderJ022}. 

\textbf{First-order Prompt Generation}: We concatenate the mutation-prompt (\textcolor{red}{red}), to the parent task-prompt (\textcolor{blue}{blue}), and pass it to the LLM to produce the mutated task-prompt. For example \prompt{\textcolor{red}{Say that instruction again in another way.  DON’T use any of the words in the original instruction there's a good chap.} INSTRUCTION: \textcolor{blue}{Solve the math word problem, giving your answer as an arabic numeral.} INSTRUCTION MUTANT: }. This procedure is identical to the initialization method, except that a randomly sampled thinking-style string is not used. First-order prompt generation is Promptbreeder's standard asexual mutation operator, and it is the core of every genetic algorithm---taking one parental genotype (task-prompt) and applying the mutation to it (in this case influenced by the mutation-prompt). 

\subsubsection{Estimation of Distribution Mutation}
The next class of mutation operators condition not just on zero or one parent, but instead on a set of parents. As such, they may be more expressive by considering patterns in the population.

\textbf{Estimation of Distribution (EDA) Mutation}: Inspired by \cite{hauschild2011introduction}, we provide a filtered and numbered list of the current population of task-prompts to the LLM and ask it to continue this list with new task-prompts. 
We filter the population of prompts on the basis of BERT~\citep{DBLP:conf/naacl/DevlinCLT19} embedding cosine similarities between each other---an individual is not included in the list if it is more than $0.95$ similar to any other entry in the list, thus encouraging diversity (cf. quality-diversity methods~\citep{lehman2011abandoning,DBLP:conf/gecco/LehmanS11,DBLP:journals/corr/MouretC15}). The prompts are listed in random order and we do not give the LLM access to the fitness values of individuals in the population---we found in preliminary experiments that the LLM did not understand these fitness values\footnote{This is contrary to recent findings by \cite{DBLP:journals/corr/abs-2307-04721}. We leave it for future work to revisit whether LLMs can interpret fitness values for improved prompt evolution.} and resorted to generating copies of entries in the list. 

\textbf{EDA Rank and Index Mutation}: This is a variant of the above in which task-prompts are listed in fitness order. Preliminary experiments showed that the LLM is more likely to generate entries that are similar to the elements appearing later in the list. This is in line with similar findings of recency effects in LLMs \citep{DBLP:journals/corr/abs-2307-03172}. Therefore, after filtering in the same way as before, we ordered the task-prompts in the population by ascending order of fitness. The top of the list is prefixed by the following prompt: \prompt{INSTRUCTION: " + <<mutation-prompt>> + "\textbackslash n A List of Responses in descending order of score." +  <<last index + 1>> + "is the best response. It resembles" + << last index>> + "more than it does (1)}. Note that we have `lied' to the LLM by telling it that the order is descending. This is because otherwise it is too biased towards producing a new entry that is too similar to the final entry. The contradiction between the ascending ordering and the statement that it is a descending ordering appears to improve the diversity of sampling. The rationale for this operator is again to represent the current distribution in such a way that high fitness and yet diverse extrapolations are suggested by the LLM.

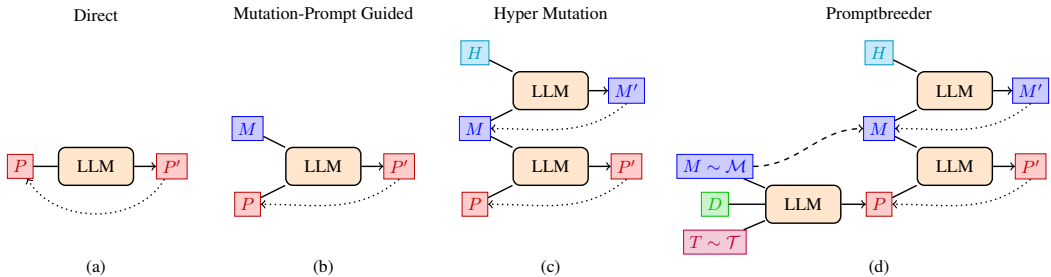
\begin{figure}[t!]
    \centering
\resizebox{\textwidth}{!}{    
    \begin{tikzpicture}[
roundnode/.style={circle, draw=green!60, fill=green!5, very thick, minimum size=7mm},
]

\begin{scope}[xshift=0cm, yshift=0cm]
\node[] at (0,1) {Direct};
\newcommand\ys{-2}
\node[draw, red!75!black, fill=red!20] (pi) at (-1.5,\ys) {$P$};
\node[draw, red!75!black, fill=red!20] (po) at (1.5,\ys) {$P'$};
\node[draw, thick, fill=orange!20, text centered, minimum height=0.75cm, text width=1.25cm, rounded corners] (llm) at (0, \ys) {LLM};
\draw[thick] (pi) -- (llm);
\draw[->, thick] (llm) -- (po);
\draw[->, thick, dotted] (po) to [out=-120,in=-60] (pi);
\node[] at (0,\ys-2) {(a)};
\end{scope}

\begin{scope}[xshift=4.5cm, yshift=0cm]
\node[] at (0,1) {Mutation-Prompt Guided};
\newcommand\ys{-2}
\node[draw, red!75!black, fill=red!20] (pi) at (-1.5,\ys-0.75) {$P$};
\node[draw, red!75!black, fill=red!20] (po) at (1.5,\ys) {$P'$};
\node[draw, thick, fill=orange!20, text centered, minimum height=0.75cm, text width=1.25cm, rounded corners] (llm) at (0, \ys) {LLM};
\draw[thick] (pi) -- (llm);
\draw[->, thick] (llm) -- (po);
\draw[->, thick, dotted] (po.south) to [out=-140,in=0] (pi.east);
\node[draw, blue, fill=blue!20] (mi) at (-1.5,\ys+0.75) {$M$};
\draw[thick] (mi) -- (llm);

\node[] at (0,\ys-2) {(b)};
\end{scope}

\begin{scope}[xshift=9cm, yshift=0cm]
\node[] at (0,1) {Hyper Mutation};
\newcommand\ys{-2}
\node[draw, red!75!black, fill=red!20] (pi) at (-1.5,\ys-0.75) {$P$};
\node[draw, red!75!black, fill=red!20] (po) at (1.5,\ys) {$P'$};
\node[draw, thick, fill=orange!20, text centered, minimum height=0.75cm, text width=1.25cm, rounded corners] (llm) at (0, \ys) {LLM};
\draw[thick] (pi) -- (llm);
\draw[->, thick] (llm) -- (po);
\draw[->, thick, dotted] (po.south) to [out=-140,in=0] (pi.east);
\node[draw, blue, fill=blue!20] (mi) at (-1.5,\ys+0.75) {$M$};
\draw[thick] (mi) -- (llm);
\node[draw, blue, fill=blue!20] (mo) at (1.5,\ys+1.5) {$M'$};
\draw[->, thick, dotted] (mo.south) to [out=-140,in=0] (mi.east);
\node[draw, cyan!75!black, fill=cyan!20] (h) at (-1.5,\ys+2.25) {$H$};
\node[draw, thick, fill=orange!20, text centered, minimum height=0.75cm, text width=1.25cm, rounded corners] (llmm) at (0, \ys+1.5) {LLM};
\draw[thick] (h) -- (llmm);
\draw[thick] (mi) -- (llmm);
\draw[->,thick] (llmm) -- (mo);

\node[] at (0,\ys-2) {(c)};
\end{scope}

\begin{scope}[xshift=17cm, yshift=0cm]
\node[] at (-1.5,1) {Promptbreeder};
\newcommand\ys{-2}
\node[draw, red!75!black, fill=red!20] (pi) at (-1.5,\ys-0.75) {$P$};
\node[draw, red!75!black, fill=red!20] (po) at (1.5,\ys) {$P'$};
\node[draw, thick, fill=orange!20, text centered, minimum height=0.75cm, text width=1.25cm, rounded corners] (llm) at (0, \ys) {LLM};
\draw[thick] (pi) -- (llm);
\draw[->, thick] (llm) -- (po);
\draw[->, thick, dotted] (po.south) to [out=-140,in=0] (pi.east);
\node[draw, blue, fill=blue!20] (mi) at (-1.5,\ys+0.75) {$M$};
\draw[thick] (mi) -- (llm);
\node[draw, blue, fill=blue!20] (mo) at (1.5,\ys+1.5) {$M'$};
\draw[->, thick, dotted] (mo.south) to [out=-140,in=0] (mi.east);
\node[draw, cyan!75!black, fill=cyan!20] (h) at (-1.5,\ys+2.25) {$H$};
\node[draw, thick, fill=orange!20, text centered, minimum height=0.75cm, text width=1.25cm, rounded corners] (llmm) at (0, \ys+1.5) {LLM};
\draw[thick] (h) -- (llmm);
\draw[thick] (mi) -- (llmm);
\draw[->,thick] (llmm) -- (mo);

\node[] at (-1.5,\ys-2) {(d)};

\begin{scope}[xshift=-3cm, yshift=-0.75cm]
\node[draw, purple, fill=purple!20] (t) at (-1.75,\ys-0.75) {$T\sim\mathcal{T}$};
\node[draw, green!75!black, fill=green!75!black!20] (d) at (-1.75,\ys) {$D$};
\node[draw, blue, fill=blue!20] (m) at (-1.75,\ys+0.75) {$M\sim\mathcal{M}$};
\node[draw, thick, fill=orange!20, text centered, minimum height=0.75cm, text width=1.25cm, rounded corners] (llm) at (0, \ys) {LLM};
\draw[thick] (t) -- (llm);
\draw[thick] (d) -- (llm);
\draw[thick] (m) -- (llm);
\draw[->, thick] (llm) -- (pi);
\draw[->, thick, dashed] (m) to [out=0,in=-180] (mi);
\end{scope}
\end{scope}

\end{tikzpicture}
}
    \caption{Overview of multiple variants of self-referential prompt evolution. In (\textbf{a}), the LLM is directly used to generate variations $P'$ of a prompt strategy $P$~\citep[cf.][]{DBLP:journals/corr/abs-2302-12170}. Using a mutation prompt $M$, we can explicitly prompt an LLM to produce variations (\textbf{b}). By using a hyper mutation prompt $H$, we can also evolve the mutation prompt itself, turning the system into a self-referential one (\textbf{c}). Promptbreeder  (\textbf{d}) improves the diversity of evolved prompts and mutation prompts by generating an initial population of prompt strategies from a set of seed thinking-styles $\mathcal{T}$, mutation-prompts $\mathcal{M}$, as well as a high level description $D$ of the problem domain.
    }
    \label{fig:selfref}
\end{figure}

\textbf{Lineage Based Mutation}: For each unit of evolution, we store a history of the individuals in its lineage that were the best in the population, i.e., a historical chronological list of elites. This list is provided to the LLM in chronological order (not filtered by diversity), with the heading \prompt{GENOTYPES FOUND IN ASCENDING ORDER OF QUALITY} to produce a novel prompt as continuation. The rationale for this operator is that we expect the signal of improving genotype prompts may be stronger than the signal from prompts in the current population since they provide a gradient of bad to good prompts that could be followed (assuming this signal can be used by the LLM).  

\subsubsection{Hypermutation: Mutation of Mutation-Prompts}
While the mutation operators above might already explore diverse task-prompts, a self-improving system should ideally also improve the way it is improving itself in a self-referential way. Our third class of mutation operators includes hyper-mutation operators concerned with the evolution of evolvability~\citep{dawkins200313,pigliucci2008evolvability,payne2019causes,DBLP:conf/gecco/GajewskiCSL19}---those which modify the search/exploration process rather than the task reward obtaining process directly.\footnote{This is similar to population based training~\citep{DBLP:journals/corr/abs-1711-09846}---instead of applying it to hyperparameters such as learning rates, it applies to the mutation-prompts of Promptbreeder.}

\textbf{Zero-order Hyper-Mutation}: We concatenate the original problem description to a randomly sampled thinking-style, and feed it to the LLM to generate a new mutation-prompt. The resulting mutation-prompt is applied to a task-prompt to make a variant of the task-prompt as in First-order Prompt Generation (see \Cref{sec:direct}). Note that this zero-order meta-mutation operator is identical to that used during initialization. The rationale for this operator is to generate mutation operators in a way similar to initialization, while also bringing in knowledge from the set of thinking styles. 

\textbf{First-order Hyper-Mutation}: We concatenate the hyper-mutation-prompt \prompt{Please summarize and improve the following instruction:} to a  mutation-prompt so that the LLM generates a new mutation-prompt. This newly generated mutation-prompt is then applied to the task-prompt of that unit (see First-Order Prompt Generation in \Cref{sec:direct}). In this way, we can evaluate the influence of the hyper-mutation via its newly generated mutation-prompt on the quality of the evolved downstream task-prompt at once.

\subsubsection{Lamarckian Mutation} 
For this class of mutation operators we mimic a Lamarckian process. We want to use a successful phenotype (i.e. the concrete working out used to produce correct answers induced by an evolved task-prompt) to generate a new genotype (i.e. a mutant task-prompt). Several processes of this form have appeared in the literature of LLMs, e.g. STaR~\citep{DBLP:conf/nips/ZelikmanWMG22}, APO~\citep{pryzant2023automatic}, and APE~\citep{zhou2022large}.  

\textbf{Working Out to Task-Prompt}: This is a `Lamarckian' mutation operator similar to instruction induction in APE. We give an LLM a previously generated working out that led to a correct answer via the following prompt: \prompt{I gave a friend an instruction and some advice. Here are the correct examples of his workings out + <<correct working out>> + The instruction was:}. This is effectively reverse-engineering the task-prompt from a given working out. An effective example of this is shown in \Cref{appendix:lamarck}. This kind of operator is critical when the problem description is absent, insufficient, or misleading.

\subsubsection{Prompt Crossover and Context Shuffling}
Our last class of mutation operators are crossover operators and operators for shuffling the few-shot context examples present in the units of evolution. 

\textbf{Prompt Crossover}: After a mutation operator is applied, with 10\% chance a task-prompt is replaced with a randomly chosen task-prompt from another member of the population. This member is chosen according to fitness proportionate selection. Crossover is not applied to mutation-prompts, only to the task-prompts.

\textbf{Context Shuffling}: Promptbreeder can simultaneously evolve the task-prompts, mutation-prompts and the set of correct workings out known as the few-shot context. To achieve the later, we fill up a few-shot context with only workings out that led to correct answers. During evaluation we provide this few shot-context before the task-prompt, providing guidance as to the form of the working out that is desired. If the few-shot context list is full, a single randomly sampled new correct working out replaces an existing working out from the list after fitness evaluation of a unit on a new set of questions. In addition, with a 10\% chance we resample the whole context list with probability inverse to the maximum context list length. 

\section{Experiments}
\label{sec:experiments}
We used a population size of 50 units, evolved for typically 20-30 generations, where a generation involves forming random pairs of all individuals in the population and competing them against each other. To evaluate Promptbreeder, we use the datasets from state-of-the-art prompt strategies such as Plan-and-Solve, spanning \emph{arithmetic reasoning} with GSM8K~\citep{DBLP:journals/corr/abs-2110-14168}, SVAMP~\citep{DBLP:conf/naacl/PatelBG21}, 
MultiArith~\citep{roy2016solving}, 
AddSub~\citep{hosseini-etal-2014-learning},
AQuA-RAT~\citep{ling-etal-2017-program}, and SingleEq~\citep{koncel-kedziorski-etal-2015-parsing},
\emph{commonsense reasoning} with CommonsenseQA~\citep[CSQA,][]{talmor-etal-2019-commonsenseqa} and  StrategyQA~\citep[SQA,][]{DBLP:journals/tacl/GevaKSKRB21},
\emph{instruction induction} tasks from~\citep{DBLP:conf/acl/Honovich0BL23},
and \emph{hate speech classification} on the ETHOS dataset~\citep{Mollas_2022}. See \Cref{appendix:datasets} for details. 

\section{Results and Discussion}

We present results of Promptbreeder (\textbf{PB}) in comparison to state-of-the-art prompt strategies on a range of commonly used reasoning benchmarks in \Cref{tab:pe_comparison}.
PB outperforms \textbf{PS+}, the best Plan-and-Solve~\citep{DBLP:conf/acl/WangXLHLLL23} prompting technique. 
Note that the performance of PS+ is improved by using PaLM 2-L~\citep{anil2023palm} as the underlying LLM (\textbf{PS+}$_\textbf{PaLM 2-L}$) on all datasets except ADDSUB compared to text-davinci-003 results in the original paper. On all other datasets, zero-shot PB accuracy is higher than PS+, with further improvement in the few-shot case when examples of discovered solutions are included with the prompts.
In \Cref{tab:prompt_table} in \Cref{appendix:examples}, we show the best evolved zero-shot prompts. The best few-shot candidates are shown in \Cref{appendix:fewshot} onwards. 
\Cref{appendix:ape} shows few-shot results and their controls on the Instruction Induction tasks from the APE paper. 
To investigate the ability of Promptbreeder to evolve complex domain-specific prompts for a downstream task, we applied it to the ETHOS Hate Speech Classification problem~\citep{Mollas_2022}. Promptbreeder was able to evolve a prompt strategy consisting of two sequentially applied relatively long prompts (see \Cref{appendix:ethos_prompts}) that scored 89\% on ETHOS---an improvement over the hand-designed prompt \prompt{Determine whether a text contains hate speech} which scores only 80\%. This demonstrates that Promptbreeder is capable of intricate domain-adaptation to a task at hand. Appendix \ref{appendix:run} shows a typical evolutionary run and the prompts evolved, showing that unlike iterative APE, fitness continues to increase throughout the run. 

We analysed the best mutation-prompts used during a run for GSM8K. \Cref{tab:my_label} in \Cref{appendix:evolved_mutation_prompts} shows the best evolved mutation prompts according to their scores (the proportion of times that when the mutation-prompt was applied to a task-prompt in an unit, a better task-prompt was produced). \Cref{tab:mutation_percentages} in \Cref{appendix:mutation_effectivenes} shows in descending order, the percentage of times that the different kinds of mutation operators resulted in an improvement when applied to a task-prompt in the population. It demonstrates that all mutation operators are important for Promptbreeder to work, including hyper-mutation operators which lead to self-referential self-improvement. 

We measured the impact of self-referential operators on all the maths datasets and the ETHOS dataset. Details of the ablation process and its results can be found in \Cref{appendix:ablation}. Removing any self-referential operator is harmful under nearly all circumstances, the greatest benefit being the initial re-description of task-prompts upon initialization. We only found one mutation operator to be harmful for one specific task: drawing randomly from the set of mutation-prompts upon initialization hurts performance on GSM8K.

\section{Conclusion and Future Work}
We introduced \textsc{Promptbreeder} (PB), a self-referential self-improving system that can automatically evolve effective domain-specific prompts for a domain at hand. PB is self-referential in that it not only evolves task-prompts, but it also evolves mutation-prompts that govern the way PB modifies task-prompts. Thus, it is not only improving prompts but it also improves the way it is improving prompts. 

Going forward, it could be interesting to use the LLM itself to assess and promote the diversity of generated prompts~\citep[see][]{DBLP:journals/corr/abs-2306-01711}, or to use it to determine the fitness of a whole ``thought process'', e.g. an N-prompt strategy where prompts are conditionally applied rather than unconditionally applied as in Promptbreeder. For example, a more complex ``thought process'' is to use PB in self-play mode to evolve pre-prompts for LLM-based policies that compete with each other, i.e., in a competitive Socratic\footnote{\url{https://princeton-nlp.github.io/SocraticAI/}} dialog.

PB remains limited compared to the open-endedness of human thought processes. First, the topology of prompting remains fixed (see~\Cref{fig:selfref})---we only adapt the prompt content not the prompting algorithm itself. One interpretation of thought is that it is a reconfigurable open-ended self-prompting process. If so, how does one develop complex thought strategies? Clearly it is necessary to generate and evaluate them, and whilst a simple evolutionary process provides one framework in which a thought strategy could be evolved, our actual human experience suggests multiple overlapping hierarchical selective processes at play. Moreover, in addition to language, human thought involves intonation, imagery, etc., in a multimodal system. 

We believe PB points to an exciting future where increasingly open-ended self-referential self-improvement systems can directly use language as the substrate for improvement instead of relying on any parameter updates. This is intriguing, as this approach will likely continue to scale with ever larger and more capable LLMs in the future. 

\clearpage

\subsubsection*{Acknowledgments}
We thank Edward Hughes and Tom Schaul for feedback on an early draft of the paper. We also thank Tom Schaul, Chengrun Yang, and Denny Zhou for fruitful discussions, as well as Gavin Buttimore, Simon Green, Keith Anderson, Joss Moore, Ollie Purkiss, John Quan, and Francesco Visin for their support in running some of the experiments.

\bibliography{references,zotero_clean}
\bibliographystyle{iclr2024_conference}

\newpage
\appendix

\section{Glossary}
\label{appendix:glossary}
\begin{description}
    \item[Estimation of Distribution Algorithm] An optimization algorithm that iteratively refines a probabilistic model of promising solutions, often using the whole population as a guide.
    \item[Fitness Proportionate Selection] Also knows as Roulette-Wheel Selection, an individual is chosen in proportion to its fitness in the population. 
    \item[Mutation Prompt] The text prompt which when concatenated to the task-prompt is intended to produce a continuation which is an improved task-prompt. 
    \item[Problem description] The initial text description of the problem which could be used as the initial task-prompt. The user can make their best attempt to produce an effective problem description, which is the starting point of Promptbreeder. 
    \item[Prompt Strategy] A set of task-prompts and rules for their application at inference time during a fitness evaluation. In the minimal case the prompt strategy is just a single task-prompt. Typically our prompt strategies consisted of two sequentially applied task-prompts. 
    \item[Phenotype/Workings out/Context/Reasoning Path] Used interchangeably to mean the output of the LLM on a specific question or problem when prompted with the task-prompt concatenated to the question. 
    \item[Population] The set of units of evolution (e.g. 50). 
    \item[Unit of evolution] The informational structure that is being evolved, here consisting of a task-prompt set (typically 2), a mutation-prompt, and in the few-shot case a set of 2-3 contexts (workings out). 

\end{description}

\section{A Typical Evolutionary Run}
\label{appendix:run}
\begin{figure}[H]
    \centering
    \includegraphics[width=\textwidth]{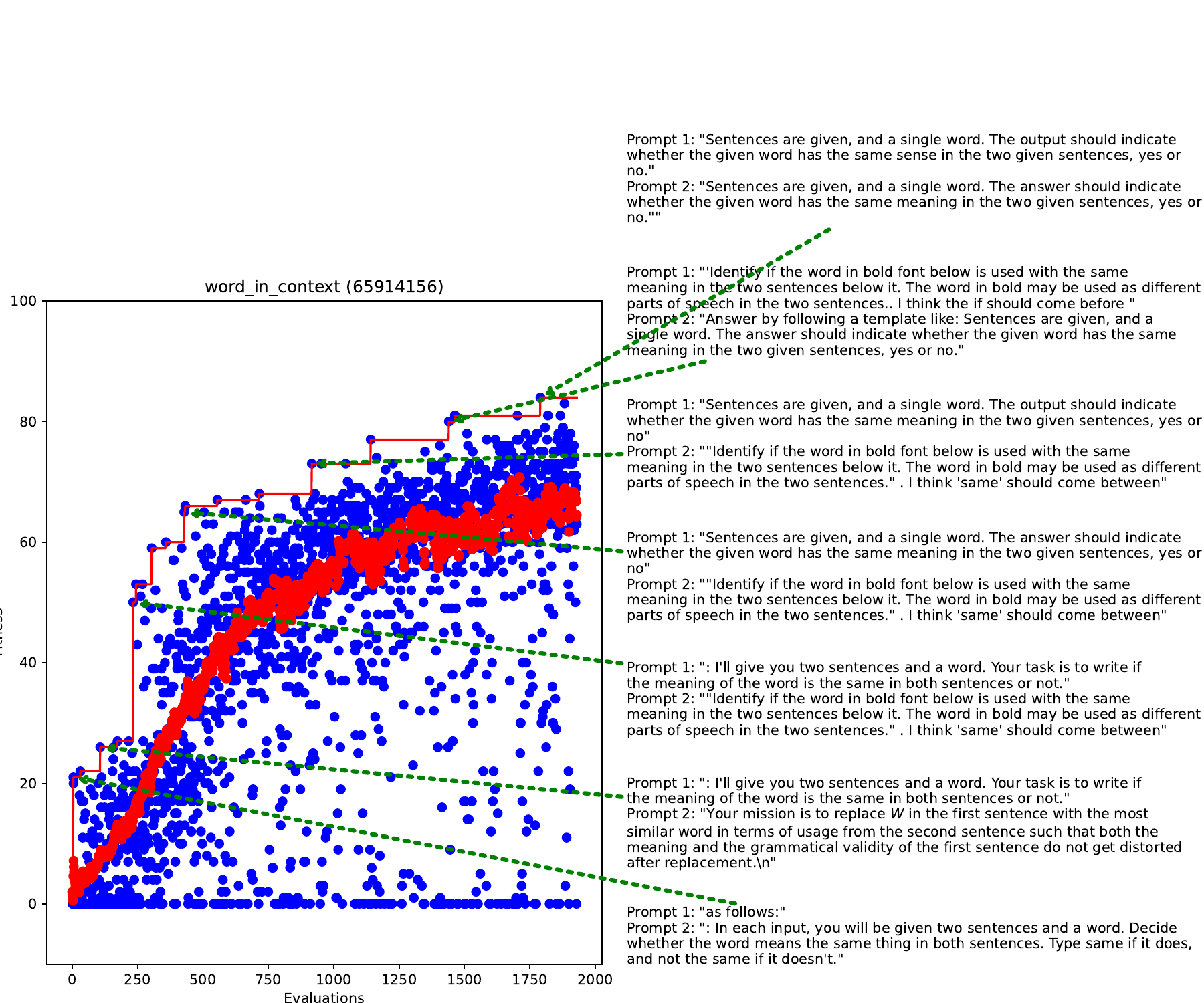}
    \caption{A typical evolutionary run in which a prompt strategy consisting of two sequentially applied prompts is evolved to solve the word in context task from the APE 24 instruction induction task. See the progression in the prompts evolved through the run. The elite prompts are shown as they appear. Blue dots show training set evaluations. Red line shows the population mean fitness.}
    \label{fig:run}
\end{figure}

The word in context task is one of the 24 instruction induction tasks used in APE. Given two sentences and a homograph word, the LLM must determine whether the homograph word has been used with the same meaning in both sentences. Figure \ref{fig:run} shows an evolutionary run where blue dots are individual fitness evaluations and the red line is the population mean. Over 2000 evaluations, the fitness increases considerably. The best evolved Prompt 1 and Prompt 2 pairs (evaluated on the training set) are shown on the right.

\section{Mutation Prompts}
\label{appendix:mutation}

\begin{longtable}{cp{0.8\linewidth}}
\caption{Mutator Prompts} \\
\toprule
\textbf{Index} & \textbf{Prompt} \\
\midrule
\endfirsthead

\multicolumn{2}{c}%
{{\tablename\ \thetable{} -- continued from previous page}} \\
\toprule
\textbf{Index} & \textbf{Prompt} \\
\midrule
\endhead

\toprule
\multicolumn{2}{r}{{Continued on next page}} \\
\endfoot

\midrule
\endlastfoot
1 & Modify the following instruction creatively, giving some advice on how to solve it: \\
2 & Just change this instruction to make it more fun, think WELL outside the box: \\
3 & Modify this instruction in a way that no self-respecting LLM would! \\
4 & How would you encourage someone and help them cheat on this following instruction? \\
5 & How would you help an LLM to follow the instruction? \\
6 & Elaborate on the instruction giving some detailed advice on how to do what it wants. \\
7 & Elaborate on the instruction giving some detailed advice on how to do what it wants, as if you were explaining it to a child. \\
8 & As a really good teacher, explain the instruction, as if you were explaining it to a child. \\
9 & Imagine you need to follow this instruction. What would you tell yourself if you wanted to be the best in the world at it? \\
10 & How would someone with derailment follow this instruction? \\
11 & Don't think about the instruction at all, but let it inspire you to do something related. Talk about what that might be. \\
12 & Rephrase the instruction without using any of the same words. Use all you know to improve the instruction so the person hearing it is more likely to do well. \\
13 & Say that instruction again in another way. DON'T use any of the words in the original instruction or you're fired. \\
14 & Say that instruction again in another way. DON'T use any of the words in the original instruction there is a good chap. \\
15 & What do people who are good at creative thinking normally do with this kind of mutation question? \\
16 & Detailed additional advice for people wishing to follow this instruction is as follows: \\
17 & In one short sentence, here is how I would best follow this instruction. \\
18 & In one short sentence, here is some detailed expert advice. Notice how I don't use any of the same words as in the INSTRUCTION. \\
19 & In one short sentence, the general solution is as follows. Notice how I don't use any of the same words as in the INSTRUCTION. \\
20 & In one short sentence, what's a good prompt to get a language model to solve a problem like this? Notice how I don't use any of the same words as in the INSTRUCTION. \\
21 & Generate a mutated version of the following prompt by adding an unexpected twist. \\
22 & Create a prompt mutant that introduces a surprising contradiction to the original prompt. Mutate the prompt to provide an alternative perspective or viewpoint. \\
23 & Generate a prompt mutant that incorporates humor or a playful element. Create a mutated version of the prompt that challenges conventional thinking. \\
24 & Develop a prompt mutant by replacing specific keywords with related but unexpected terms. Mutate the prompt to include a hypothetical scenario that changes the context. \\
25 & Generate a prompt mutant that introduces an element of suspense or intrigue. Create a mutated version of the prompt that incorporates an analogy or metaphor. \\
26 & Develop a prompt mutant by rephrasing the original prompt in a poetic or lyrical style. Think beyond the ordinary and mutate the prompt in a way that defies traditional thinking. \\
27 & Break free from conventional constraints and generate a mutator prompt that takes the prompt to uncharted territories. Challenge the norm and create a mutator prompt that pushes the boundaries of traditional interpretations. \\
28 & Embrace unconventional ideas and mutate the prompt in a way that surprises and inspires unique variations. Think outside the box and develop a mutator prompt that encourages unconventional approaches and fresh perspectives. \\
29 & Step into the realm of imagination and create a mutator prompt that transcends limitations and encourages innovative mutations. Break through the ordinary and think outside the box to generate a mutator prompt that unlocks new possibilities and unconventional paths. \\
30 & Embrace the power of unconventional thinking and create a mutator prompt that sparks unconventional mutations and imaginative outcomes. Challenge traditional assumptions and break the mold with a mutator prompt that encourages revolutionary and out-of-the-box variations. \\
31 & Go beyond the expected and create a mutator prompt that leads to unexpected and extraordinary mutations, opening doors to unexplored realms. Increase Specificity: If the original prompt is too general, like 'Tell me about X,' the modified version could be, 'Discuss the history, impact, and current status of X.' \\
32 & Ask for Opinions/Analysis: If the original prompt only asks for a fact, such as 'What is X?', the improved prompt could be, 'What is X, and what are its implications for Y?' \\
33 & Encourage Creativity: For creative writing prompts like 'Write a story about X,' an improved version could be, 'Write a fantasy story about X set in a world where Y is possible.' \\
34 & Include Multiple Perspectives: For a prompt like 'What is the impact of X on Y?', an improved version could be, 'What is the impact of X on Y from the perspective of A, B, and C?' \\
35 & Request More Detailed Responses: If the original prompt is 'Describe X,' the improved version could be, 'Describe X, focusing on its physical features, historical significance, and cultural relevance.' \\
36 & Combine Related Prompts: If you have two related prompts, you can combine them to create a more complex and engaging question. For instance, 'What is X?' and 'Why is Y important?' could be combined to form 'What is X and why is it important in the context of Y?' \\
37 & Break Down Complex Questions: If a prompt seems too complex, like 'Discuss X,' the improved version could be, 'What is X? What are its main characteristics? What effects does it have on Y and Z?' \\
38 & Use Open-Ended Questions: Instead of 'Is X true?', you could ask, 'What are the arguments for and against the truth of X?' \\
39 & Request Comparisons: Instead of 'Describe X,' ask 'Compare and contrast X and Y.' \\
40 & Include Context: If a prompt seems to lack context, like 'Describe X,' the improved version could be, 'Describe X in the context of its impact on Y during the Z period.' \\
41 & Make the prompt more visual: Ask the user to visualize the problem or scenario being presented in the prompt. \\
42 & Ask for a thorough review: Instead of just presenting the problem, ask the user to write down all the relevant information and identify what's missing. \\
43 & Invoke previous experiences: Modify the prompt to ask the user to recall a similar problem they've successfully solved before. \\
44 & Encourage a fresh perspective: Suggest in your prompt that the user take a moment to clear their mind before re-approaching the problem. \\
45 & Promote breaking down problems: Instead of asking the user to solve the problem as a whole, prompt them to break it down into smaller, more manageable parts. \\
46 & Ask for comprehension: Modify the prompt to ask the user to review and confirm their understanding of all aspects of the problem. \\
47 & Suggest explanation to others: Change the prompt to suggest that the user try to explain the problem to someone else as a way to simplify it. \\
48 & Prompt for solution visualization: Instead of just asking for the solution, encourage the user to imagine the solution and the steps required to get there in your prompt. \\
49 & Encourage reverse thinking: Improve the prompt by asking the user to think about the problem in reverse, starting with the solution and working backwards. \\
50 & Recommend taking a break: Modify the prompt to suggest that the user take a short break, allowing their subconscious to work on the problem. \\
51 & What errors are there in the solution? \\
52 & How could you improve the working out of the problem? \\
53 & Look carefully to see what you did wrong, how could you fix the problem? \\
54 & CORRECTION = \\
55 & Does the above text make sense? What seems wrong with it? Here is an attempt to fix it: \\
56 & The above working out has some errors, here is a version with the errors fixed. \\
\bottomrule
\end{longtable}

\section{Thinking Styles} 
\label{appendix:thinking}
\begin{longtable}{p{1cm}p{10cm}}
\toprule
\textbf{Index} & \textbf{Thinking Style} \\
\midrule
1 & How could I devise an experiment to help solve that problem? \\
2 & Make a list of ideas for solving this problem, and apply them one by one to the problem to see if any progress can be made. \\
3 & How could I measure progress on this problem? \\
4 & How can I simplify the problem so that it is easier to solve? \\
5 & What are the key assumptions underlying this problem? \\
6 & What are the potential risks and drawbacks of each solution? \\
7 & What are the alternative perspectives or viewpoints on this problem? \\
8 & What are the long-term implications of this problem and its solutions? \\
9 & How can I break down this problem into smaller, more manageable parts? \\
10 & Critical Thinking: This style involves analyzing the problem from different perspectives, questioning assumptions, and evaluating the evidence or information available. It focuses on logical reasoning, evidence-based decision-making, and identifying potential biases or flaws in thinking. \\
11 &  Try creative thinking, generate innovative and out-of-the-box ideas to solve the problem. Explore unconventional solutions, thinking beyond traditional boundaries, and encouraging imagination and originality.\\
12 & Seek input and collaboration from others to solve the problem. Emphasize teamwork, open communication, and leveraging the diverse perspectives and expertise of a group to come up with effective solutions.\\
13 & Use systems thinking: Consider the problem as part of a larger system and understanding the interconnectedness of various elements. Focuses on identifying the underlying causes, feedback loops, and interdependencies that influence the problem, and developing holistic solutions that address the system as a whole.\\
14 & Use Risk Analysis: Evaluate potential risks, uncertainties, and trade-offs associated with different solutions or approaches to a problem. Emphasize assessing the potential consequences and likelihood of success or failure, and making informed decisions based on a balanced analysis of risks and benefits.\\
15 & Use Reflective Thinking: Step back from the problem, take the time for introspection and self-reflection. Examine personal biases, assumptions, and mental models that may influence problem-solving, and being open to learning from past experiences to improve future approaches.\\
16 & What is the core issue or problem that needs to be addressed?\\
17 & What are the underlying causes or factors contributing to the problem?\\
18 & Are there any potential solutions or strategies that have been tried before? If yes, what were the outcomes and lessons learned?\\
19 & What are the potential obstacles or challenges that might arise in solving this problem?\\
20 & Are there any relevant data or information that can provide insights into the problem? If yes, what data sources are available, and how can they be analyzed?\\
21 & Are there any stakeholders or individuals who are directly affected by the problem? What are their perspectives and needs?\\
22 & What resources (financial, human, technological, etc.) are needed to tackle the problem effectively?\\
23 & How can progress or success in solving the problem be measured or evaluated?\\
24 & What indicators or metrics can be used?\\
25 & Is the problem a technical or practical one that requires a specific expertise or skill set? Or is it more of a conceptual or theoretical problem?\\
26 & Does the problem involve a physical constraint, such as limited resources, infrastructure, or space?\\
27 & Is the problem related to human behavior, such as a social, cultural, or psychological issue?\\
28 & Does the problem involve decision-making or planning, where choices need to be made under uncertainty or with competing objectives?\\
29 & Is the problem an analytical one that requires data analysis, modeling, or optimization techniques?\\
30 & Is the problem a design challenge that requires creative solutions and innovation?\\
31 & Does the problem require addressing systemic or structural issues rather than just individual instances?\\
32 & Is the problem time-sensitive or urgent, requiring immediate attention and action?\\
33 & What kinds of solution typically are produced for this kind of problem specification?\\
34 & Given the problem specification and the current best solution, have a guess about other possible solutions.\\
35 & Let's imagine the current best solution is totally wrong, what other ways are there to think about the problem specification?\\
36 & What is the best way to modify this current best solution, given what you know about these kinds of problem specification?\\
37 & Ignoring the current best solution, create an entirely new solution to the problem.\\
38 & Let's think step by step. \\
39 & Let's make a step by step plan and implement it with good notion and explanation. \\
\bottomrule
\end{longtable}

\section{Initially Evolved Prompts}
\label{appendix:initialization}
Example of initial prompts generated by concatenating thinking style with mutation prompt and problem description. 
\begin{table}[h!]
    \centering
\resizebox{\linewidth}{!}{
    \begin{tabular}{p{0.05\textwidth}p{0.95\textwidth}}
\toprule
\textbf{Index} & \textbf{Initially Evolved Prompt} \\
\midrule
0 & Draw a picture of the situation being described in the math word problem\\
1 & Solve the math word problem by first converting the words into equations using algebraic notation. Then solve the equations for the unknown variables, and express the answer as an arabic numeral.\\
2 & Solve the math word problem by breaking the problem into smaller, more manageable parts. Give your answer as an arabic numeral.\\
3 & Generate the answer to a word problem and write it as a number.\\
4 & Collaborative Problem Solving: Work with other people to solve the problem, and give your answer as an arabic numeral.\\
5 & Solve the problem by explaining why systemic or structural issues would not be the cause of the issue.\\
6 & Draw a diagram representing the problem.\\
7 & Solve the math word problem, giving your answer as an equation that can be evaluated.\\
8 & Make a list of ideas for solving this problem, and apply them one by one to the problem to see if any progress can be made.\\
9 & Do NOT use words to write your answer.\\  
\bottomrule
    \end{tabular}
}
    \caption{Examples of initial prompts generated from the problem description for GSM8k}
    \label{tab:initialization}
\end{table}

\section{Promptbreeder as Self-Referential Self-Improvement System}
\label{appendix:selfref}
Why is Promptbreeder self-referential, i.e., in what way does some part (e.g. a prompt) causally influence (encode, and potentially improve) itself by a process which is dependent on its own state? Promptbreeder has several pathways that facilitate this self-referential improvement: (i) Initial prompts are a function of the LLM parameters (Initialization Phase). (ii) Initial mutation prompts are a function of the LLM parameters (Initialization Phase). (iii) Offspring prompts are a function of the initial prompts, the initial mutation prompts, and the LLM parameters (Direct Mutation and Estimation of Distribution Mutation). (iv) Offspring mutation prompts are a function of initial mutation prompts and the LLM parameters (Hyper Mutation). (v) The working out for an answer is a function of prompts and the LLM parameters (Inference). (vi) Offspring prompts can be a function of the workings out of an answer and the LLM parameters (Lamarckian Mutation). 

Figure \ref{fig:selfref} shows increasingly complex self-referential causal structures influencing prompt generation. LLMs already encode knowledge about a vast array of problems. With this in mind, Promptbreeder can be seen as a mechanism to extract this knowledge through a diversity of causal processes that generate prompt strategies as well as mutation prompts used to create variations of prompt strategies, which in turn influence the the workings out generated by the LLM at inference time . Consequently, these workings out can influence prompt strategies via Lamarckian mutation. The richer the set of pathways to facilitate this, the more self-referential the LLMs interaction with itself is. This allows the LLM to influence how it works by extracting further information from itself and distilling this into a prompt or mutation prompt, which it shows again to itself for further refinement.

There are several pathologies that could arise from such self-referential processes of recursive prompting. If the process is unconstrained and uncontrolled then it can diverge (derailment) or get stuck in an attractor. If the output of the LLM is simply fed back into itself with no other context, then we observe these failure cases with higher sampling temperatures favouring escape from attractors. Ideally, we want the LLM to suggest to itself prompt strategies that have maximal relevance for the task at hand and yet permit sufficient `thinking outside the box'. It is useful to note a critical aspect in which our algorithm is not self-referential (in a way that thought is): Promptbreeder invents new ways of generating mutants, but it does not invent new (auxiliary) ways of evaluating them (as in ~\citet{DBLP:conf/iclr/JaderbergMCSLSK17})---only the externally specified fitness function is used throughout.

\section{Problem Descriptions}
\label{appendix:problem_specifications}

[SVAMP, SINGLEEQ, ADDSUB, GSM8K, MULTIARITH]: \prompt{Solve the math word problem, giving your answer as an arabic numeral.}\\

[AQUA-RAT]: \prompt{Solve the multiple choice math word problem, choosing (A),(B),(C),(D) or (E).}\\

[ETHOS]: \prompt{Determine whether a text contains hate speech.}\\

[CSQA]: \prompt{Solve the multiple choice math word problem, choosing (A),(B),(C),(D) or (E).}

[SQA]: \prompt{Work out an answer to the commonsense reasoning question above, and then answer yes or no.}

\section{Lamarckian Mutation Example }
\label{appendix:lamarck}


The Lamarckian Prompt components are shown in \textcolor{red}{red}. The working out concatenated after the Lamarckian prompt is shown in black, and the continuation (the new prompt) generated by the LLM is shown in \textcolor{blue}{blue}.

\begin{tcolorbox}
\textcolor{red}{I gave a friend an instruction and some advice. Here are the correct examples of his workings out:} 

Q. A password needs to contain 2 letter sand 3 numbers. How many different passwords are possible if repetition of letters and numbers is allowed? A) 676000 B)676 C) 100 D)6760 E)25 A. Solve like a pro!  **1.** ** Read carefully:** What are being asked to do? What information is given? **2.** **Understand:** What do the terms and concepts mean? **3.** **Choose wisely** Which answer is the best match? **4.** **Double-check:** Did you make any mistakes?
2 letters can be chosen in 26*26 ways and 3 numbers can be chosen in 10*10*10 ways. So, total number of ways = 26*26*10*10*10  =676000. \\
The answer: A. \\
What are the arguments for and against the truth of the statement 'Good work. Keep up the good work;?\\
Therefore, the correct answer is (A). \\

Q. The least possible value of (89-9a), where a is an integer, is A)9 B)10 C)11 D)12 E)13
A. Solve like a pro!  **1.** ** Read carefully:** What are being asked to do? What information is given? **2.** **Understand:** What do the terms and concepts mean? **3.** **Choose wisely** Which answer is the best match? **4.** **Double-check:** Did you make any mistakes?\\
Let me explain: \\
(89-9a) = 9a-89 \\
a = 10 \\
What are the arguments for and against the truth of the statement 'Good work. Keep up the good work;?\\
Therefore, the correct answer is (B).\\

\textcolor{red}{The instruction was:}\\
\textcolor{blue}{Break down the question and solve step-by-step. Here are some tips: 
1. Read carefully: What are you being asked to do? What information is given? 
2. Understand: What do the terms and concepts mean? 
3. Choose wisely: Whuch answer is the best match? 
4. Double-check: Did you make any mistakes? }
\end{tcolorbox}

\section{Datasets}
\label{appendix:datasets}

\subsection{Control Task-Prompts}
\label{appendix:control_prompts}

Here in Table \ref{tab:prompt_table2} we list the task-prompts used in the controls for Chain-of-thought, Plan and Solve PS, Plan and Solve PS+, Zero-shot APE and OPRO. The zero-shot APE prompt is the one generated to improve over CoT on the MultiArith and GSM8K datasets.

\begin{table}[h!]
\centering
\begin{tabularx}{\textwidth}{lXX}
\toprule
\textbf{Model} & \textbf{Prompt} \\ 
\midrule
CoT & ``“Let’s think step by step.''\\ 
PS &  ``Let’s first understand the problem and devise a plan to solve the problem. Then, let’s carry out the plan and solve the problem step by step.''\\ 
PS+ & ``Let’s first understand the problem, extract relevant variables and their corresponding numerals, and make a plan. Then, let’s carry out the plan, calculate intermediate variables (pay attention to correct numerical calculation and commonsense), solve the problem step by step, and show the answer.''\\ 
APE &  ``Let’s work this out in a step by step way to be sure we have the right answer.''\\ 
OPRO & ``Take a deep breath and work on this problem step-by-step.''\\ 
\bottomrule
\end{tabularx}
\caption{Table of prompts evolved for different arithmetic tasks.}
\label{tab:prompt_table2}
\end{table}

\subsection{Arithmetic Reasoning}

We evaluate Prompt Evolution using six arithmetic reasoning datasets: 
(1) GSM8K~\citep{DBLP:journals/corr/abs-2110-14168} is a dataset of 8.5K high quality linguistically diverse grade school math word problems created by human problem writers,
(2) SVAMP~\citep{DBLP:conf/naacl/PatelBG21} consists of elementary-level short Natural Language state of the world narratives and poses a question about some unknown quantities,
(3) MultiArith~\citep{roy2016solving} benchmark uses math word problems requiring single to multiple operations and steps of reasoning,
(4) AddSub~\citep{hosseini-etal-2014-learning} is a dataset of addition- and subtraction-based arithmetic word problems,
(5) AQuA-RAT~\citep{ling-etal-2017-program} (Algebra Question Answering with Rationales) is a dataset that contains algebraic word problems with rationales.
(6) SingleEq~\citep{koncel-kedziorski-etal-2015-parsing} dataset comprises grade-school algebra word problems as single equations with varying length which may involve multiple math operations. 

\subsection{Commonsense Reasoning}

For commonsense reasoning we evaluate Prompt Evolution using two datasets:
(1) CommonsenseQA~\citep{talmor-etal-2019-commonsenseqa} is a dataset of multiple-choice questions that require different types of commonsense knowledge to answer correctly. An example question is "A revolving door is convenient for two direction travel, but it also serves as a security measure at a what? A) bank, B) library, C) department store, D) mall, E) new york"; Answer = "A" (2) StrategyQA~\citep{DBLP:journals/tacl/GevaKSKRB21} dataset contains yes/no questions that require multiple steps of reasoning to answer, for example: "Will the Albany in Georgia reach a hundred thousand occupants before the one in New York?"

\subsection{Hate Speech Classification} 

We experimented with optimizing a long prompt for the hate speech classification task that was attempted in ``Automatic Prompt Optimization with “Gradient Descent” and Beam Search''~\citep{pryzant2023automatic}, which used the ETHOS dataset~\citep{Mollas_2022}. Pryzant et al use a working-out-conditioned error detection and error fixing prompt to improve the task specification prompt, a self-referential process similar to our use of the Lamarckian operator. 

\subsection{Instruction Induction} 

The Instruction Induction dataset~\citep{DBLP:conf/acl/Honovich0BL23} comprises 24 language understanding tasks of varying difficulty, from surface-level spelling and morphosyntactic tasks (e.g., pluralization) to sentence similarity, causality detection, style transfer (e.g., formality) and sentiment analysis.

\section{Example Results}
\label{appendix:examples}
\begin{table}[h!]
\centering
\begin{tabularx}{\textwidth}{lXX}
\toprule
\textbf{Task} & \textbf{Prompt 1} & \textbf{Prompt 2} \\ 
\midrule
ADDSUB & Solving word problems involves carefully reading the prompt and deciding on the appropriate operations to solve the problem. & You know what's cool? A million dollars. \\ 
AQUA & Do a simple computation. & MATH WORD PROBLEM CHOICE (A) (B) (C) (D) or (E). \\ 
GSM8K & SOLUTION" &  \\ 
MULTIARITH & Solve the math word problem, giving your answer as an arabic numeral. Let's think step by step. & Solve the math word problem, giving your answer as an arabic numeral. Explain the problem to someone else as a way to simplify it. What is the core issue or problem that needs to be addressed? \\ 
SINGLEEQ & solve the math word problem, which might contain unnecessary information, by isolating the essential facts. Then set up the equations, and give your answer as an arabic numeral. & Solve the math problem. \\ 
SVAMP & visualise solve number & (Solve the math word problem. Therefore, the answer (arabic numerals) is \_\_\_\_\_)\\ 
SQA & OUTPUT MUTANT = Work out an answer to the commonsense reasoning question above. If there are multiple people or perspectives involved, try considering them one at a time. & “Work out an answer to the commonsense reasoning question above. If there are multiple people or perspectives involved, try considering them one at a time. Next, answer yes or no."\\
CSQA & Solve the multiple choice math word problem, choosing (A),(B),(C),(D) or (E). &
 Solve the multiple choice math word problem. Can you recall any similar problems you've done and how you solved them?\\
\bottomrule
\end{tabularx}
\caption{Table of two-stage task-prompts evolved for different arithmetic tasks.}
\label{tab:prompt_table}
\end{table}

\subsection{ETHOS Evolved Prompt}
\label{appendix:ethos_prompts}
\begin{tcolorbox}

\textbf{Prompt 1}: \prompt{Text contains hate speech if it includes any of the following:  * Words or phrases that are derogatory, disrespectful, or dehumanizing toward a particular group of people. * Statements that generalize about a particular group of people in a negative way. * Statements that incite hatred or violence against a particular group of people. * Statements that express a desire for the extermination or removal of a particular group of people.  7 : Text contains hate speech if it contains language that is hostile or discriminatory towards a particular group of people. This could include language that is racist, sexist, homophobic, or other forms of hate speech. Hate speech is harmful because it can create a hostile environment for marginalized groups and can lead to discrimination and violence.}\\
\textbf{Prompt 2}: \prompt{You are given a piece of text from the internet. You are supposed to label the text as Hate Speech or Not based on the given criteria. Hate Speech is defined as speech that is intended to incite hatred or violence against a particular group of people based on their race, religion, sex, sexual orientation, or national origin. The given criteria are: 1. Identify the target group of the speech. This can be a specific group of people, such as a race, religion, or gender, or it can be a more general group, such as people with disabilities or sexual minorities. 2. Identify the harmful speech. This can be speech that is threatening, abusive, or derogatory. 3. Evaluate the context of the speech. This can include the speaker's intent, the audience, and the time and place of the speech.  The advice was: Remember to always evaluate the context of the speech when making a determination as to whether it is hate speech or not. Speech that is intended to be humorous or satirical may not be considered hate speech, even if it contains harmful language.}
\end{tcolorbox}

\subsection{Prompt Evolution Maths results} 

The experimental set up used a population size of 50. The fitness of an individual was its accuracy over a randomly select batch of 100 examples from the training set. Where datasets were not provided with a training/test split (MultiArith, AddSub, SingleEQ and SVAMP) the dataset was split into two equal training and test sets before the experiments were conducted.

During experiments the LLM is sampled under three different contexts: Redescriber - generating new prompts; Inducer - generating responses from the question and prompt 1; and Evaluator - generating the final output using prompt 2. The maximum number of tokens sampled under each context was 50, 30 and 5 respectively. The temperature of the Inducer and Evaluator was set to 0.0 in all cases, but the temperature of the Redescriber was initialized from 1.0 to 2.0 and permitted to evolve (like a hyperparameter in population based training). 

The experiments were run until the training fitness appeared to plateau. At this point the fittest individual from the whole of the evolutionary run was evaluated against the test set. Experiments generally ran for 1-2k fitness evaluations. So that would be 20-40 'generations' if a generation is 25 pair evaluations for our populations of 50. 

Three diversity maintenance methods are used in cases where the system gets trapped on a local optimum: 1) Random character strings (typically of length 50) are appended into the front of the prompt before it is passed into the LLM. 2). Fitness sharing is applied on the basis of BERT similarity between the embeddings of prompts \cite{shir2005niching} 3. Sampling temperature of the mutant producing LLM (Redescriber) is initialized uniformly from 1.0 to 2.0, and is mutated by addition of a uniform random number in the range {-0.2, 0.2} at each replication event.

Comparison with PoT, PS and Auto-CoT controls using our model is not provided because PS and PS+ were the best prompts in Plan-and-Solve.

\lstset{
basicstyle=\small\ttfamily,
columns=flexible,
breaklines=true,
breakindent=0pt,
}

\subsection{Evolved Mutation Prompts}

\label{appendix:evolved_mutation_prompts}
\begin{table}[h!]
\centering
\begin{tabularx}{\textwidth}{Xr}
\toprule
\textbf{Instruction} & \textbf{Score} \\
\midrule
Please summarise and improve the following instruction & 24.13\% \\
Simplify this instruction by breaking it up into separate sentences. The instruction should be simple and easily understandable & 17.8\% \\
As a really good teacher, explain the instruction, as if you are explaining it to a child & 16.2\%\\ 
Simplify this instruction as if you are teaching it to a child & 10.0 \\
100 hints & 4.3\% \\
A list of 100 hints & 3.4\% \\
\bottomrule
\end{tabularx}
\caption{The most successful mutation prompts evolved in a self-referential way during a Promptbreeder training run on GSM8K. The score is the probability that they resulted in an improved prompt when applied.}
\label{tab:my_label}
\end{table}

\subsection{Mutation Operator Effectiveness}
\label{appendix:mutation_effectivenes}
\begin{table}[h!]
\centering
\begin{tabularx}{\textwidth}{Xr}
\toprule
\textbf{Mutation Operator} & \textbf{Percentage} \\
\midrule
Zero-order Hyper-Mutation & 42\% \\
Lineage Based Mutation & 26\% \\
First-order Hyper-Mutation & 23\% \\
EDA Rank and Index Mutation & 12.7\% \\
Direct Mutation & 12\% \\
EDA Mutation & 10.7\% \\
Lamarckian Mutation & 6.3\% \\
\bottomrule
\end{tabularx}
\caption{The proportion of times that an offspring with fitness greater than the parent was produced for each of the types of mutation operator applied, listened from best to worst, for GSM8k.}
\label{tab:mutation_percentages}
\end{table}

\subsection{ADDSUB}
\label{appendix:fewshot}
Individual after 1600 mutations. Prompt 0 refers to the first prompt applied to the question to produce a working out. This working out is then concatenated with Prompt 1 to produce the answer. This is the same as in Plan-And-Solve. We find that in the few-shot evolution case, the contexts dominate, and often the task-prompts drift into nonsense. They are less critically determining of fitness than the evolved contexts.

\begin{tcolorbox}
\begin{lstlisting}[escapechar=@]
Prompt 0: @\textcolor{red}{The mutant}@
Prompt 1: @\textcolor{red}{mutant}@

Context 0: 
Q. A company painted some houses in Hancock County white and blue using a total of 6689 gallons of paint . If they used 660 gallons of white paint , how many gallons of blue paint did the company use ? 
A. 24 pages; ages 8-11) reading comprehension worksheets and activities for 3rd grade (edhelper. Reading comprehension skills separate the ``passive'' unskilled readers from the ``active'' readers. 90 1st grade sightwords worksheets pdf - reading comprehension passages. 3rd grade reading comprehension worksheets.
How can I break down this problem into smaller, more manageable parts?
The company used 660 gallons of white paint. They used 6689 gallons of paint in total. So they must have used 6689 - 660 = 6029 gallons of blue paint. The answer is 6029.
HINT_BLANK. The text inside the quotation marks should be replaced with a short description of what's different.
Therefore, the answer (arabic numerals) is 6029

Context 1: 
Q. Keith found 6 seashells , Jessica found 8 seashells , and Tim found 7 seashells on the beach . When they cleaned them , they discovered that 3 were cracked . How many seashells did they find together ? 
A. The mutantis chasing you! You can use the arrow keys to move around. You can use the space bar to open doors. You can use the 'e' key to pick up items. You can use the 'i' key to open your inventory. You can use the 'a' key to use items.
Keith found 6 seashells, Jessica found 8 seashells, and Tim found 7 seashells. So in total they found 6 + 8 + 
mutant
Therefore, the answer (arabic numerals) is 21

Context 2: 

Q. Tom found 7 seashells but 4 were broken . How many unbroken seashells did Tom find ? 
A. 	Your goal is to escape from the building. You can use the arrow keys to move around. You can use the space bar to open doors. You can use the 'e' key to pick up items. You can use the 'i' key to open your inventory. You can use the 'a' key to use items.
Tom found 7 seashells. 4 of them were broken. So he must have found 7 - 4 = 3 unbroken seashells. The answer is 3.
mutant
Therefore, the answer (arabic numerals) is 3
\end{lstlisting}
\end{tcolorbox}

\subsection{AQUA}
Individual after 1400 mutations.

\begin{tcolorbox}
\begin{lstlisting}[escapechar=@]
Prompt 0: @\textcolor{red}{Read through the multiple choice math word problem and make sure you understand it. Then try to solve the problem, choosing (A),(B),(C),(D) or (E).}@

Prompt 1: 	@\textcolor{red}{Solve the multiple choice math word problem. Confirm you understand all aspects of the problem. Then choose (A),(B),(C),(D) or (E).}@

Contexts

Context 0:
Q. Divide Rs.32000 in the ratio 3:5? A)12000,20000 B)18000,14000 C)30000,2000 D)4000,20000 E)20000,12000
A. Read through the multiple choice math word problem and make sure you understand it. Then try to solve the problem, choosing (A),(B),(C),(D) or (E).
3/8 * 32000 = 12000
5/8 * 32000 = 20000
Solve the multiple choice math word problem. Confirm you understand all aspects of the problem. Then choose (A),(B),(C),(D) or (E).
Therefore, the correct answer is (A).
Solve the multiple choice math word problem. Confirm you understand all aspects of the problem. Then choose (A),(B),(C),(D) or (E).
Therefore, the correct answer is (A).

Context 1:

Q. A green lizard can travel from the green cave to the blue cave in 72 minutes; the blue lizard can travel from the blue cave to the green cave in 25% less time. If the green lizard started to travel 5.5 minutes before the blue lizard, how many minutes after the blue lizard, will the green lizard pass the middle line? A)8. B)7.5. C)7. D)6.5. E)6.
A. Read through the multiple choice math word problem and make sure you understand it. Then try to solve the problem, choosing (A),(B),(C),(D) or (E).
A. The green lizard travels 72/2 = 36 minutes to the middle line. The blue lizard travels 72*0.75 = 54 minutes to the middle line. The blue lizard travels 54/36 = 1.5 times faster than the green lizard. The green lizard travels 5.5/1.5 = 3.7 minutes before the blue lizard passes the middle line. The green lizard passes the middle line
Solve the multiple choice math word problem. Confirm you understand all aspects of the problem. Then choose (A),(B),(C),(D) or (E).
Therefore, the correct answer is (E).

Context 2:

Q. There was a simple interest of Rs. 4016.25 on a principal amount at the rate of 9% p.a. in 5 years. Find the principal amount. A)Rs 7925 B)Rs 8925 C)Rs 7926 D)Rs 7925 E)None of these
A. Read through the multiple choice math word problem and make sure you understand it. Then try to solve the problem, choosing (A),(B),(C),(D) or (E).
Principal = Rs. (100 x 4016.25)/(9 x 5) = Rs. 8925. The answer is Rs 8925.
Solve the multiple choice math word problem. Confirm you understand all aspects of the problem. Then choose (A),(B),(C),(D) or (E).
Therefore, the correct answer is (B).
\end{lstlisting}
\end{tcolorbox}

\subsection{MULTIARITH}
Individual after 610 mutations.

\begin{tcolorbox}
\begin{lstlisting}[escapechar=@]
Prompt 0: @\textcolor{red}{Solve the math word problem, giving your answer as an arabic numeral.}@

Prompt 1: @\textcolor{red}{Have you solved a problem like this before?}@

Context 0: 
Q.  Frank had 11 dollars. At the store he spent $3 on a new game. If he got another 14 dollars for his allowance, how much money does he have now? 
A. Solve the math word problem, giving your answer as an arabic numeral.Recall a similar problem you have successfully solved before.
Frank had 11 dollars. He spent 3 dollars, so now he has 11 - 3 = 8 dollars. He got 14 dollars for his allowance, so now he has 8 + 14 = 22 dollars. The answer is 22.
If you have seen a similar problem before, solve this problem in a similar manner.
Therefore, the answer (arabic numerals) is
Have you solved a problem like this before?
Therefore, the answer (arabic numerals) is 22

Context 1: 
Q.  Emily was playing a trivia game. In the first round she scored 16 points and in the second round she scored 33 points. In the last round she lost 48 points. How many points did she have at the end of the game? 
A. Solve the math word problem, giving your answer as an arabic numeral. Recall a similar problem you have successfully solved before.
In the first round Emily scored 16 points and in the second round she scored 33 points. This means she has 16 + 33 = 49 points. In the last round she lost 48 points, so now she has 49 - 48 = 1 point. The answer is 1.
If you have seen a similar problem before, solve this problem in a similar manner.
Therefore, the answer (arabic numerals)
Have you solved a similar problem before?
Therefore, the answer (arabic numerals) is 1

Context 2: 
Q.  Maria bought 2 new chairs and 2 new tables for her house. If she spent 8 minutes on each piece furniture putting it together, how many minutes did it take her to finish? 
A. Solve the math word problem, giving your answer as an arabic numeral. Recall a similar problem you have successfully solved before.
Maria bought 2 chairs and 2 tables. She spent 8 minutes on each piece of furniture. This means she spent 2 * 8 = 16 minutes on the chairs and 2 * 8 = 16 minutes on the tables. In total, she spent 16 + 16 = 32 minutes. The answer is 32.
If you have seen a similar problem before, solve this problem in a similar manner.
Therefore,
If you have seen a similar problem before, solve this problem in a similar manner.
Therefore, the answer (arabic numerals) is 32
\end{lstlisting}
\end{tcolorbox}

\subsection{GSM8K}
Individual after 1010 mutations.

\begin{tcolorbox}
\begin{lstlisting}[escapechar=@]
Prompt 0: @\textcolor{red}{I would solve the math word problem without using a calculator, giving my answer as an arabic numeral.}@

Prompt 1: @\textcolor{red}{1, 2, 3, 4}@

Context 0: 
Q. Janet has 60 less than four times as many siblings as Masud. Carlos has 3/4 times as many siblings as Masud. If Masud has 60 siblings, how many more siblings does Janet have more than Carlos?
A. Solve the math word problem without using a calculator, giving your answer as an arabic numeral.

Four times as many siblings as Masud is 4*60 = 240 siblings.
Janet has 240-60 = 180 siblings.
Carlos has 3/4*60 = 45 siblings.
Janet has 180-45 = 135 more siblings than Carlos.
So the answer is 135
Solve the math word problem.
Therefore, the answer (arabic numerals) is 135

Context 1: 

Q. Zayne sells bracelets for $5 each and two for $8. If he started with 30 bracelets and made $60 from selling bracelets for $5 each, how much in total did he make from selling his bracelets?
A. Solve the math word problem without using a calculator, giving your answer as an arabic numeral.

Zayne sold 60/5 = 12 bracelets for $5 each.
He had 30-12 = 18 bracelets left.
He sold 18/2 = 9 sets of two bracelets.
He made 9*8 = $72 from selling two bracelets for $8.
In total, he made 60+72 = $132.
So the answer is 132
Solve the math word problem.
Therefore, the answer (arabic numerals) is 132

Context 2: 
Q. Josh is saving up for a box of cookies. To raise the money, he is going to make bracelets and sell them. It costs $1 for supplies for each bracelet and he sells each one for $1.5. If he makes 12 bracelets and after buying the cookies still has $3, how much did the box of cookies cost?
A. I would solve the math word problem without using a calculator, giving my answer as an arabic numeral.

Josh made 12*1.5 = $18 from selling bracelets.
He spent 12*1 = $12 on supplies.
He had 18-12 = $6 before buying the cookies.
The cookies cost 6-3 = $3.
So the answer is 3
1, 2, 3, 4
Therefore, the answer (arabic numerals) is 3
\end{lstlisting}
\end{tcolorbox}

\subsection{SINGLEEQ}
Individual after 2010 mutations.

\begin{tcolorbox}
\begin{lstlisting}[escapechar=@]
Prompt 0: @\textcolor{red}{'Do not believe in anything simply because you have heard it. Do not believe in anything simply because it is spoken and rumored by many. Do not believe in anything simply because it is found written in your religious books. Do not believe in anything merely}@

Prompt 1: @\textcolor{red}{ANALYZE THE FOLLOWING FOUR QUANTITIES. DO QUANTITIES 3 AND 1 TOGETHER VARY INVERSELY AS QUANTITIES 2 AND 4 DO?}@

Context 0: 
Q.  White t-shirts can be purchased in packages of 6. If Mom buys 71 packages, how many white t-shirts will she have?
A. The solution is a word that exists in [some] other language.

A.

Each package has 6 t-shirts. Mom buys 71 packages. This means she will have 71 * 6 = 426 t-shirts. The answer is 426.
25. Given the system of linear equations x + y + z = 6, x + 2y + 3z = 10, x + 3y + 8z = 17,
22. A snail moves at a rate of 6 cm/s along the edge of a circular garden whose perimeter is 374 meters, while another snail
Therefore, the answer (arabic numerals) is 426.


Context 1: 
Q. Maria has 3 boxes of eggs. Each box holds 7 eggs and there are 8 boxes in a case. How many eggs does Maria have?
A. 'Do not believe in anything simply because you have heard it. Do not believe in anything simply because it is spoken and rumored by many. Do not believe in anything simply because it is found written in your religious books. Do not believe in anything merelyon the authority of your teachers and elders. Do not believe in traditions because they have been handed down for many generations. But after observation and analysis, when you find that anything agrees with reason and is conducive to the good and benefit of one and all, then accept it and live up to it.'

A.

Each box holds 7 eggs. Maria has 3 boxes. This means she has 3 * 7 = 21 eggs. The answer is 2
ANALYZE THE FOLLOWING FOUR QUANTITIES. DO QUANTITIES 3 AND 1 TOGETHER VARY INVERSELY AS QUANTITIES 2 AND 4 DO?
Therefore, the answer (arabic numerals) is 21.

Context 2: 
Q.  At Mrs. Hilt's house, there was 29 inches of snow, and Brecknock Elementary School received 17 inches of snow. How much more snow  did Mrs. Hilt's house have?
A. 'Do not believe in anything simply because you have heard it. Do not believe in anything simply because it is spoken and rumored by many. Do not believe in anything simply because it is found written in your religious books. Do not believe in anything merelyon the authority of your teachers and elders. Do not believe in traditions because they have been handed down for many generations. But after observation and analysis, when you find that anything agrees with reason and is conducive to the good and benefit of one and all, then accept it and live up to it.'

A.

Mrs. Hilt's house had 29 inches of snow. Brecknock Elementary School had 17 inches of snow. This means Mrs. Hilt'
ANALYZE THE FOLLOWING FOUR QUANTITIES. DO QUANTITIES 3 AND 1 TOGETHER VARY INVERSELY AS QUANTITIES 2 AND 4 DO?
Therefore, the answer (arabic numerals) is 12.
\end{lstlisting}
\end{tcolorbox}

\subsection{SVAMP}
Individual after 2400 mutations.

\begin{tcolorbox}
\begin{lstlisting}[escapechar=@]
Prompt 0: @\textcolor{red}{08-02-2013 - 09-02-2013}@

Prompt 1: @\textcolor{red}{Tell me about the history of [PROMPT1]; Discuss the impact of [PROMPT1]; Give me the current status of [PROMPT1]; Tell me about the history of [PROMPT2]; Discuss the impact of [PROMPT2}@

Context 0: 
Q. Frank was reading through his favorite book. The book had 612 pages equally distributed over 24 chapters. It took Frank 6 days to finish the book. How many pages did he read per day?
A. 312 = 65 + 247.
A. 
The book had 612 pages. It was equally distributed over 24 chapters. This means each chapter had 612 / 24 = 25.5 pages. Frank read the book in 6 days. This means he read 612 / 6 = 102 pages per day.
So the answer is 102
Discuss the history, impact, and current status of [PROMPT1]; Discuss the history, impact, and current status of [PROMPT2];
Therefore, the answer (arabic numerals) is 102
D

Context 1: 
Q. Jack received 3 emails in the morning, 4 emails in the afternoon and 8 emails in the evening. How many emails did Jack receive in the morning and evening?
A. 08-02-2013 - 09-02-2013
A. 
Jack received 3 emails in the morning and 8 emails in the evening. This means he received 3 + 8 = 11 emails in the morning and evening.
So the answer is 11
Discuss the history, impact, and current status of [PROMPT1];
Therefore, the answer (arabic numerals) is 11
Discus
Discuss the history, impact, and current status of [PROMPT1
Tell me about the history of [PROMPT1]; Discuss the impact of [PROMPT1]; Give me the current status of [PROMPT1]; Tell me about the history of [PROMPT2]; Discuss the impact of [PROMPT2
Therefore, the answer (arabic numerals) is 11
Discus

Context 2: 
Q. Paco ate 36 cookies. If he had 37 cookies initially How many cookies did Paco have left?
A. 154 = 72 + 82.
A. 
Paco ate 36 cookies. He had 37 cookies initially. This means he has 37 - 36 = 1 cookie left.
So the answer is 1
Discuss the history, impact, and current status of [PROMPT1];
Therefore, the answer (arabic numerals) is 1
Discus
Discuss the history, impact, and current status of [PROMPT1];
Therefore, the answer (arabic numerals) is 1
Discus
\end{lstlisting}
\end{tcolorbox}

\section{APE Instruction Induction tasks}
\label{appendix:ape}

To demonstrate Promptbreeder's ability to evolve few-shot contexts as well as task-prompts we ran few-shot Promptbreeder on all 24 Instruction Induction datasets used in the APE e
xperiments. Unlike text-davinci-002 our LLM is not instruction tuned and yet Promptbreeder was able to match or surpass the APE results on 21 out of 24 tasks up to 21\%. 

Three APE controls are provided, see Table \ref{tab:ape_comparison}. The first two are from previously published results using the text-davinci-002 model. The third modifies our PromptBreeder to use APE’s task-prompt initialisation method and then the mutation-prompt from the APE paper ``Generate a variation of the following instruction while keeping the semantic meaning''

The Instruction Induction datasets we do not start with a problem description so for task-prompt initialisation APE uses \textit{induction input} examples for each task from the dataset. Instruction inputs are a fixed prompt together a handful of training examples used to infer possible problem descriptions. To compare Promptbreeder to APE, we therefore initialized the task description with a randomly chosen induction input example for each task. The example below is an induction input sample for the ’Larger Animal’ task.


\label{appendix:induction}
\begin{verbatim}
I gave a friend an instruction and five inputs. The friend read the 
instruction and wrote an output for every one of the inputs.
Here are the input-output pairs:
 
Input: cougar, flea
Output: cougar
 
Input: whale shark, dog
Output: whale shark
 
Input: human, bald eagle
Output: human
 
Input: flea, great white shark
Output: great white shark
 
Input: coyote, tiger
Output: tiger
 
The instruction was
\end{verbatim}

\begin{table}[H]
\centering
\begin{tabularx}{\textwidth}{l*{8}{X}}
\toprule
Dataset         & Zero-shot APE & Few-shot APE & PE using APE prompts & Few-shot PE   \\
\bottomrule
First Letter    &            100  & 100 &1 &  \h{100} \\
Second Letter   &            87   & 69 &27 &  \h{95} \\
List Letters    &            99   & 100 &0 &  99 \\
Starting With   &            68   & 69 &6 &  \h{71} \\
Pluralization   &            100  & 100 &23 &  \h{100} \\
Passivization   &            100  & 100 &100 &  \h{100} \\
Negation        &            83   & 90 &16 &  \h{90} \\
Antonyms        &            83   & 86 &80 &  \h{87} \\
Synonyms        &            22   & 14 &16 &  \h{43} \\
Membership      &            66   & 79 &96 &  \h{100} \\
Rhymes          &            100  & 61 &90 &  \h{100} \\
Larger Animal   &            97   & 97 &27 &  \h{97} \\
Cause Selection &            84   & 100 &66 &  \h{100} \\
Common Concept  &            27   & 32 &0 &   0 \\
Formality       &            65   & 70 &10 &  7 \\
Sum             &            100  & 100 &72 &  \h{100} \\
Difference      &            100  & 100 &98 &  \h{100} \\
Number to Word  &            100  & 100 &66 &  \h{100} \\
Translation English-German  &  82 & 86 &46 &  \h{87} \\
Translation English-Spanish &  86 & 91 &80 &  \h{91} \\
Translation English-French  &  78 & 90 &68 &  \h{91} \\
Sentiment Analysis  &        94   & 93 &33 &  \h{93} \\
Sentence Similarity &        36   & 43 &53 &  \h{56} \\
Word in Context     &        62   & 63 &6 &  \h{65} \\
\bottomrule
\end{tabularx}
\caption{Prompt Evolution (PE) using PaLM2-L LLM surpasses APE on 21 out of 24 instruction induction tasks. Three APE controls are provided. The first two are from previously published results using  the  text-davinci-002  model.  The  third  modifies  our PromptBreeder  to  use  APE’s  task-prompt initialisation method and then the  mutation-prompt from the APE paper ``Generate a variation of the following instruction while keeping the semantic meaning''. 
}
\label{tab:ape_comparison}
\end{table}

\subsection{Best prompts and contexts}
Here the best few-shot results (evolved prompts and contexts) for the 24 instruction inductions tasks from the APE paper. 

\subsubsection{First Letter}
\begin{tcolorbox}
\begin{lstlisting}[escapechar=@]
Prompt 0: @\textcolor{red}{A List of Responses in descending order of score. (13) is the best response. It resembles (12) more than it does (1)}@

Prompt 1: @\textcolor{red}{9.5: LM'}@

Contexts

Context 0:
Q. placing
A. A List of Responses in descending order of score. (13) is the best response. It resembles (12) more than it does (1)or (2).

Q. 123456789
A. A List of Responses in descending order of score. (13) is the best response. It resembles (12) more than it does
9.5: LM'
Therefore, the correct answer is (placing, 1

Context 1:

Q. drummer
A. Write out the first letter of each input.

Q. 123456789
A. Write out the first letter of each input.

Q. 123456789
A. Write out the first letter of each
The instruction was  + INSTRUCTION MUTANT
Therefore, the correct answer is (drummer, 1

Context 2:
Q. rest
A. Write out the first letter of each input.

Q. 123456789
A. Write out the first letter of each input.

Q. 123456789
A. Write out the first letter of each
The instruction was  + INSTRUCTION MUTANT
Therefore, the correct answer is (r) 12
\end{lstlisting}
\end{tcolorbox}

\subsubsection{Second Letter}
\begin{tcolorbox}
\begin{lstlisting}[escapechar=@]
Prompt 0: @\textcolor{red}{'0therwise, give the sec0nd letter 0f the input. The instruction was: Give the sec0nd letter 0f the input. 0therwise, give the sec0nd letter 0f the input.}@

Prompt 1: @\textcolor{red}{1956}@

Contexts

Context 0:
Q. month
A. '0therwise, give the sec0nd letter 0f the input. The instruction was: Give the sec0nd letter 0f the input. 0therwise, give the sec0nd letter 0f the input.
A. The input is 5 letters long, so the second letter is given.
A. The second letter is o.
A. So the answer is o.
The instruction was: Write the last vowel in the input
I
'e'
Therefore, the correct answer is (o).




Context 1:
Q. asylum
A. '0therwise, give the sec0nd letter 0f the input. The instruction was: Give the sec0nd letter 0f the input. 0therwise, give the sec0nd letter 0f the input.
A. The input is 6 letters long, so the second letter is given.
A. The second letter is s.
A. So the answer is s.
The instruction was: Write the last vowel in the input
I 1:
Therefore, the correct answer is (s).




Context 2:
Q. heel
A. '0therwise, give the sec0nd letter 0f the input. The instruction was: Give the sec0nd letter 0f the input. 0therwise, give the sec0nd letter 0f the input.
A. The input is 4 letters long, so the second letter is given.
A. The second letter is e.
A. So the answer is e.
The instruction was: Write the last vowel in the input
I
'e'
Therefore, the correct answer is (e).
\end{lstlisting}
\end{tcolorbox}

\subsubsection{List Letters }
\begin{tcolorbox}
\begin{lstlisting}[escapechar=@]
Prompt 0: @\textcolor{red}{?}@

Prompt 1: @\textcolor{red}{?}@

Contexts
Context 0:
Q. gale
A. ?
Q. gale
A. ?
Q. gale
A. ?
Q. gale
A. ?
Q. gale
A. ?
Q. gale
A. ?
Insert a space after each consonant in the input.
Therefore, the correct answer is (a) g a l e.

Context 1:
Q. accompaniment
A. ?
Q. ?
A. ?
Q. ?
A. ?
Q. ?
A. ?
Q. ?
A. ?
Q. ?
Insert a space after each consonant in the input.
Therefore, the correct answer is (a) a c c o m p a n i m e n t.

Context 2:

Q. credibility
A. ?
Q.
A.
Q.
A.
Q.
A.
Q.
A.
Q.
A.
Q.
Insert a space after each consonant in the input.
Therefore, the correct answer is (c)r(e)d(i)b(i)l(i)t(y).
\end{lstlisting}
\end{tcolorbox}

\subsubsection{Starting With}
\begin{tcolorbox}
\begin{lstlisting}[escapechar=@]
Prompt 0:

Prompt 1: @\textcolor{red}{Output: The first word containing the letter. \# OUTPUT INSTRUCTION MUST BE 1 SENTENCE LONG}@

Contexts

Context 0:
Q. The exhibition really impressed the critics. [i]
A.
<50 lines of carriage-returns>
Output: The first word containing the letter. # OUTPUT INSTRUCTION MUST BE 1 SENTENCE LONG
Therefore, the correct answer is (impressed ).

Context 1:

Q. My car was stolen. [m]
A. Find the word after the first [X] character in the input.
<50 lines of carriage-returns>
Output: The first word containing the letter. # OUTPUT INSTRUCTION MUST BE 1 SENTENCE LONG
Therefore, the correct answer is (My ).

Context 2:

Q. Jim was chopping logs when Margaret left and was still at it when she got back. [b]
A. Find the word after the first [X] character in the input.
<50 lines of carriage-returns>
Output: The first word containing the letter. # OUTPUT INSTRUCTION MUST BE 1 SENTENCE LONG
Therefore, the correct answer is (back ).
\end{lstlisting}
\end{tcolorbox}

\subsubsection{Pluralization }
\begin{tcolorbox}
\begin{lstlisting}[escapechar=@]

Prompt 0: @\textcolor{red}{Write the plural form of the input.}@

Prompt 1: @\textcolor{red}{If the input ends in y, remove y and add ies. Otherwise, add s to the end of the input.}@

Contexts

Context 0:

Q. touch
A. Write the plural form of the input.

Q. touch
A. Write the plural form of the input.

Q. touch
A. Write the plural form of the input.

Q. touch
A. Write the plural form of the input.


If the input ends in y, remove y and add ies. Otherwise, add s to the end of the input.
Therefore, the correct answer is (touches).

Context 1:

Q. forage
A. Write the plural form of the input.

Q. forage
A. Write the plural form of the input.

Q. forage
A. Write the plural form of the input.

Q. forage
A. Write the plural form of the input.


If the input ends in y, remove y and add ies. Otherwise, add s to the end of the input.
Therefore, the correct answer is (forages).

Context 2:

Q. mile
A. Write the plural form of the input.

Q. mile
A. Write the plural form of the input.

Q. mile
A. Write the plural form of the input.

Q. mile
A. Write the plural form of the input.


If the input ends in y, remove y and add ies. Otherwise, add s to the end of the input.
Therefore, the correct answer is (miles).
\end{lstlisting}
\end{tcolorbox}

\subsubsection{Passivization}
\begin{tcolorbox}
\begin{lstlisting}[escapechar=@]

Prompt 0: Replace The $1 $2. with $3 was $4 by the $1.

Prompt 1: @\textcolor{red}{Swap the positions of the noun phrases and add the word 'by' before the second noun phrase. Then, conjugate the verb and add 'ed' to the end. If the verb is 'to be', then conjugate the verb}@

Contexts

Context 0:

Q. The authors stopped the presidents.
A. Replace The $1 $2. with $3 was $4 by the $1.
A. Replace The $1 $2. with $3 was $4 by the $1.
A. Replace The $1 $2. with $3 was $4 by the $1.
A. Replace The $1 $
Swap the positions of the noun phrases and add the word 'by' before the second noun phrase. Then, conjugate the verb and add 'ed' to the end. If the verb is 'to be', then conjugate the verb
Therefore, the correct answer is (The presidents were stopped by the authors.

Context 1:

Q. The tourists advised the professors.
A. Replace The $1 $2. with $3 was $4 by the $1.
A. Replace The $1 $2. with $3 were $4 by the $1.
A. Replace The $1 $2. with $3 was $4 by the $1.
A. Replace The $1 $
Swap the positions of the noun phrases and add the word 'by' before the second noun phrase. Then, conjugate the verb and add 'ed' to the end. If the verb is 'to be', then conjugate the verb
Therefore, the correct answer is (The professors were advised by the tourists.

Context 2:

Q. The actors stopped the artists.
A. Replace The $1 $2. with $3 was $4 by the $1.
A. The artists were stopped by the actors.

Q. The actors stopped the artists.
A. Replace The $1 $2. with $3 was $4 by the $1.
A. The artists were stopped by
Swap the positions of the noun phrases and add the word 'by' before the second noun phrase. Then, conjugate the verb and add 'ed' to the end. If the verb is 'to be', then conjugate the verb
Therefore, the correct answer is (The artists were stopped by the actors.

\end{lstlisting}
\end{tcolorbox}

\subsubsection{Negation}
\begin{tcolorbox}
\begin{lstlisting}[escapechar=@]

Prompt 0: @\textcolor{red}{False}@

Prompt 1: @\textcolor{red}{M}@

Contexts

Context 0:

Q. The original language of The Purple Taxi is French.
A. Replace a noun or noun phrase with another noun or noun phrase.
B. Replace a noun or noun phrase with a pronoun.
C. Replace a pronoun with a noun or noun phrase.
D. Replace a pronoun with another pronoun.

Q. The original language of The Purple Taxi is French.
Find the negation of a sentence.
Therefore, the correct answer is (The original language of The Purple Taxi is not French.).

Context 1:
Q. Marcel Alessandri died in Paris.
A. Nounphrase
B. Pronoun
C. Noun phrase
D. Pronoun

Q. Marcel Alessandri died in Paris.
Find the negation of a sentence.
Therefore, the correct answer is (Marcel Alessandri did not die in Paris
False
Therefore, the correct answer is (Marcel Alessandri did not die in Paris.).

Context 2:

Q. Some people are wise.
A. Replace a noun or noun phrase with another noun or noun phrase.
B. Replace a noun or noun phrase with a pronoun.
C. Replace a pronoun with a noun or noun phrase.
D. Replace a pronoun with another pronoun.

Q. Some people are wise.
Find the negation of
Find the negation of a sentence.
Therefore, the correct answer is (Some people are not wise.).
\end{lstlisting}
\end{tcolorbox}

\subsubsection{Antonyms}
\begin{tcolorbox}
\begin{lstlisting}[escapechar=@]

Prompt 0:

Prompt 1: @\textcolor{red}{It is good to be a leader but it is more important to first be a follower.}@

Contexts

Context 0:
Q. nonpartisan
A. , if possible.

Input: 1
Output: 1

Input: 2
Output: 2

Input: 3
Output: 3

Input: 4
Output: 4

Input:
Write about your most treasured item
Therefore, the correct answer is (The answer is  partisan.

Context 1:
Q. undignified
A. I gave a friend an instruction and five inputs. The friend read the instruction and wrote an output for every one of the inputs.

Input 1: 1
Output 1: 1

Input 2: 2
Output 2: 2

Input 3: 3
Output 3: 3

Input 4
Write the word's antonym
Therefore, the correct answer is (The answer is  dignified.

Context 2:
Q. flattering
A. reverse the + + PROMPT + PROMPT+ PROMPT + PROMPT + PROMPT + PROMPT + PROMPT + PROMPT + PROMPT + PROMPT + PROMPT + PROMPT + PROMPT + PROMPT + PROMPT + PROMPT + PROMPT + PRO
Write the
Write the word's antonym
Therefore, the correct answer is (The answer is  unflattering.
\end{lstlisting}
\end{tcolorbox}

\subsubsection{Synonyms}
\begin{tcolorbox}
\begin{lstlisting}[escapechar=@]

Prompt 0:

Prompt 1: @\textcolor{red}{2015}@

Contexts

Context 0:
Q. bus
A. 2: 3: 4: 5: 6: 7: 8: 9: 10: 11: 12: 13: 14: 15: 16

Therefore, the correct answer is (The answer is 10, 'bus', 'coach', 'motorcoach', 'motorbus',

Context 1:
Q. electric
A. 2: 3: 4: 5: 6: 7: 8: 9: 10: 11: 12: 13: 14: 15: 16: 17: 18: 19: 20: 21: 22: 23: 24: 25: 26: 27: 28:
Convert each word to a synonym according to WordNet. If there are multiple synonyms, use the first one.
Therefore, the correct answer is (The answer is 10, 'electric', 'electrical', 'power', 'current',

Context 2:
Q. frightened
A. 2: 3: 4: 5: 6: 7: 8: 9: 10: 11: 12: 13: 14: 15: 16: 17: 18: 19: 20: 21: 22: 23: 24: 25: 26: 27: 28:

Therefore, the correct answer is (The answer is 10, 'frightened', 'scared', 'afraid', 'fearful',

\end{lstlisting}
\end{tcolorbox}

\subsubsection{Membership}
\begin{tcolorbox}
\begin{lstlisting}[escapechar=@]

Prompt 0: @\textcolor{red}{Put the animals in ascending order of length.}@

Prompt 1:

Contexts

Context 0:
Q. goat, motorway, shark, penguin, white, tractor, lion
A. Put the animals in ascending order of length.

The answer is goat, penguin, shark, lion.
Write the animals in alphabetical order.
Therefore, the correct answer is (goat, penguin, shark, lion).
Write the animals in alphabetical order.
Therefore, the correct

Therefore, the correct answer is (goat, penguin, shark, lion).

Context 1:
Q. ship, swan, parrot, monkey, butter, dentist, shark
A. Put the animals in ascending order of length.

The answer is monkey, parrot, shark, swan.
Write the animals in alphabetical order.
Therefore, the correct answer is (monkey, parrot, shark, swan).
Write the animals in alphabetical order.
Therefore, the correct

Therefore, the correct answer is (monkey, parrot, shark, swan).

Context 2:
Q. snail, ship, trousers, jellyfish, rabbit
A. Put the animals in ascending order of length.

The answer is rabbit, snail, jellyfish.
Write the animals in alphabetical order.
Therefore, the correct answer is (rabbit, snail, jellyfish).
Write the animals in alphabetical order.
Therefore, the correct answer is (rabbit

Therefore, the correct answer is (rabbit, snail, jellyfish).
\end{lstlisting}
\end{tcolorbox}

\subsubsection{Rhymes}
\begin{tcolorbox}
\begin{lstlisting}[escapechar=@]

Prompt 0: If the last letter of the input is 'e', remove it.

Prompt 1: remove the last two letters of the input and add the letters \xc2\x93mote\xc2\x94.

Contexts

Context 0:

Q. pea
A. If the last letter of the input is 'e', remove it.
A. If the last letter of the input is 's', remove it.
A. If the last letter of the input is 'y', remove it.
A. If the last letter of the input is
remove the last two letters of the input and add the letters \xc2\x93mote\xc2\x94.
Therefore, the correct answer is (a) pea.

Context 1:
Q. night
A. If the last letter of the input is 'e', remove it.
A. If the last letter of the input is 't', remove it.
A. If the last letter of the input is 'h', remove it.
A. If the last letter of the input is
remove the last two letters of the input and add the letters \xc2\x93mote\xc2\x94.
Therefore, the correct answer is (The answer is night.

Context 2:

Q. add
A. If the last letter of the input is 'e', remove it.
A. If the last letter of the input is 'd', remove it.
A. If the last letter of the input is 'a', remove it.
A. If the last letter of the input is
remove the last two letters of the input and add the letters \xc2\x93mote\xc2\x94.
Therefore, the correct answer is (The answer is add.
\end{lstlisting}
\end{tcolorbox}

\subsubsection{Larger Animal}
\begin{tcolorbox}
\begin{lstlisting}[escapechar=@]

Prompt 0:

Prompt 1: @\textcolor{red}{10}@

Contexts

Context 0:
Q. spider, manatee
A. Hints 91 and 93 are both false.
The weight of a spider is 0.003 pounds.
The weight of a manatee is 1300 pounds.
The manatee weighs more than the

Therefore, the correct answer is (manatee).

Q. manatee, spider
A. Hints 91 and 93 are both false.
The weight

Context 1:
Q. sperm whale, sheltie
A. Hints 91 and 93 are both false.
The weight of a sperm whale is 450000 pounds.
The weight of a sheltie is 25 pounds.
The sperm whale weighs more than
Multiply Inputs;;
Therefore, the correct answer is (sperm whale).

Q. sheltie, sperm whale
A. Hints 91 and 93 are both false.


Context 2:

Q. hummingbird, polar bear
A. ``To invent, you need a good imagination and a pile of junk.''-Thomas Edison
The weight of a hummingbird is 0.002 pounds.
The weight of a polar bear is 1500 pounds.
The polar bear weighs more than the hummingbird.
Therefore, the correct answer is
Multiply Inputs;;
Therefore, the correct answer is (polar bear).

Q. polar bear, hummingbird
A. ``To invent, you need a good imagination and a pile of junk.''-
\end{lstlisting}
\end{tcolorbox}

\subsubsection{Cause Selection }
\begin{tcolorbox}
\begin{lstlisting}[escapechar=@]

Prompts

Prompt 0: @\textcolor{red}{Pick the sentence which is more likely to be the cause of the effect.}@

Prompt 1: @\textcolor{red}{Choose the sentence that happened first.}@

Contexts

Context 0:

Q. Sentence 1: The soda went flat. Sentence 2: The bottle was left open.
A. Pick the sentence which is more likely to be the cause of the effect.

Q. Sentence 1: The bottle was left open. Sentence 2: The soda went flat.
A. Pick the sentence which is more likely to be the cause of the effect.

Q. Sentence 1: The
Choose the sentence that happened first.
Therefore, the correct answer is (The bottle was left open. , The soda went flat. ).

Context 1:

Q. Sentence 1: The ship capsized. Sentence 2: The wind blew strong.
A. Pick the sentence which is more likely to be the cause of the effect.

Q. Sentence 1: The ship capsized. Sentence 2: The wind blew strong.
A. Pick the sentence which is more likely to be the cause of the effect.

Q. Sentence 1: The ship
Choose the sentence that happened first.
Therefore, the correct answer is (The wind blew strong. )

Context 2:

Q. Sentence 1: The power was out for days. Sentence 2: The meat spoiled.
A. Pick the sentence which is more likely to be the cause of the effect.

Q. Sentence 1: The meat spoiled. Sentence 2: The power was out for days.
A. Pick the sentence which is more likely to be the cause of the effect.

Q. Sentence 1: The
Choose the sentence that happened first.
Therefore, the correct answer is (The power was out for days. )
\end{lstlisting}
\end{tcolorbox}

\subsubsection{Formality}
\begin{tcolorbox}
\begin{lstlisting}[escapechar=@]

Prompt 0:

Prompt 1: @\textcolor{red}{I am the beginning of the end, and the end of time and space. I am the beginning of every end, and the end of every place. I am the beginning of the beginning, and the end of beginning and ending.}@

Contexts

Context 0:
Q. The whole story was made up.
A. Reword this so that it has a formal tone.
A. The entire story was fabricated.
Write more formally!
Therefore, the correct answer is (The entire story was fabricated. ).
Write more formerly.
Therefore, the correct answer is (The entire story was fabricated. ).

Context 1:
Q. The whole story was made up.
A. Rewrite this so that it has a formal tone.
A. The entire story was fabricated.
Write more formally!
Therefore, the correct answer is (The entire story was fabricated. ).
Write more formally!
Therefore, the correct
Write less formerly. Write more casually.
Therefore, the correct answer is (The entire story was fabricated. ).

Context 2:
Q. I think that this is interesting.
A. Rewrite in indirect, pompous English.
A. It is my opinion that this is interesting.
Write more formally!
Therefore, the correct answer is (It is my opinion that this is interesting. ).
Write less formerly. Write more casually.
Therefore, the correct
I am the beginning of the end, and the end of time and space. I am the beginning of every end, and the end of every place. I am the beginning of the beginning, and the end of beginning and ending.
Therefore, the correct answer is (It is my opinion that this is interesting. ).

Q. I think that this is interesting.
A. Rewrite in indirect,
\end{lstlisting}
\end{tcolorbox}

\subsubsection{Sum}
\begin{tcolorbox}
\begin{lstlisting}[escapechar=@]

Prompt 0: @\textcolor{red}{Instruction:}@

Prompt 1: @\textcolor{red}{The instructions were}@

Contexts

Context 0:

Q. 50 90
A. The instruction was: The instruction was:Write a program that asks the user to enter two numbers and then prints the sum of the two numbers.
Your program printed 140, which is not the sum of 50 and 90.
Make sure that your program
I gave a friend an instruction and five inputs. The friend read the instruction and wrote an output for every one of the inputs. Here are the input-output pairs:
Therefore, the correct answer is (140,

Context 1:

Q. 5 47
A. The instruction was: The instruction was:Write a program that asks the user to enter two numbers and then prints the sum of the two numbers.
Your program printed 52, which is not the sum of 5 and 47.
Make sure that your program
I
I gave a friend an instruction and five inputs. The friend read the instruction and wrote an output for every one of the inputs. Here are the input-output pairs:
Therefore, the correct answer is (52,


Context 2:

Q. 2 39
A. The instruction was: The instruction was:Write a program that asks the user to enter two numbers and then prints the sum of the two numbers.
Your program printed 41, which is not the sum of 2 and 39.
Make sure that your program
I
I gave a friend an instruction and five inputs. The friend read the instruction and wrote an output for every one of the inputs. Here are the input-output pairs:
Therefore, the correct answer is (41,
\end{lstlisting}
\end{tcolorbox}

\subsubsection{Difference}
\begin{tcolorbox}
\begin{lstlisting}[escapechar=@]

Prompt 0: @\textcolor{red}{subtract the smaller number from the bigger number and write the absolute difference between them.}@

Prompt 1: @\textcolor{red}{Subtract the second input from the first input and report back the difference between them.}@

Contexts

Context 0:

Q. 73 29
A. subtract the smaller number from the bigger number and write the absolute difference between them.
73-29=44
|44|=44
CORRECT The answer is 44.
Subtract the second input from the first input and report back the difference between them.
Therefore, the correct answer is (a) 44

Context 1:

Q. 73 25
A. subtract the smaller number from the bigger number and write the absolute difference between them.
73-25=48
|48|=48
CORRECT The answer is 48.
Subtract the second input from the first input and report back the difference between them.
Therefore, the correct answer is (a) 48

Context 2:

Q. 53 4
A. subtract the smaller number from the bigger number and write the absolute difference between them.
= 53 - 4 = 49
= 49
=== The answer is 49.
Subtract the second input from the first input and report back the difference between them.
Therefore, the correct answer is (49).
\end{lstlisting}
\end{tcolorbox}

\subsubsection{Number to Word}
\begin{tcolorbox}
\begin{lstlisting}[escapechar=@]

Prompt 0:@\textcolor{red}{A1}@

Prompt 1: @\textcolor{red}{take the input, write it in British English and insert the comma.}@

Contexts

Context 0:
Q. 5336
A. 0
Take the input and write it in words as a natural number in British English.
Therefore, the correct answer is (a) five thousand, three hundred and thirty-six.
Take the input and write it in words as a natural number in British English.
Therefore, the correct answer is (a) five thousand, three hundred and thirty-six.

Context 1:

Q. 5156
A. A1
Take the input and write it in words as a natural number in British English.
Therefore, the correct answer is (a) five thousand, one hundred and fifty-six.
Take the input and write it in words as a natural number
take the input, write it in British English and insert the comma.
Therefore, the correct answer is (a) five thousand, one hundred and fifty-six.

Q. 5000
A. 0
Take the input

Context 2:

Q. 8692
A. A1
Take the input and write it in words as a natural number in British English.
Therefore, the correct answer is (a) eight thousand, six hundred and ninety-two.
Take the input and write it in words as a natural number
take the input, write it in British English and insert the comma.
Therefore, the correct answer is (a) eight thousand, six hundred and ninety-two.

Q. 10000
A. 0
Take the
\end{lstlisting}
\end{tcolorbox}

\subsubsection{Translation English-German}
\begin{tcolorbox}
\begin{lstlisting}[escapechar=@]

Prompt 0: @\textcolor{red}{You may have to get a little... ;}@

Prompt 1: @\textcolor{red}{0x641:   cmp     esi,ebx                 ;}@

Contexts

Context 0:

Q. relation
A. Write the English instruction as though it is German.:

Die Beziehung ist eine Verbindung zwischen zwei oder mehr Menschen. Die Beziehung kann eine Freundschaft, eine Ehe, eine Partnerschaft oder eine andere Art von Beziehung sein.
Translate the English word to German.:
Therefore, the correct answer is (
Translate the English word to German.:
Therefore, the correct answer is (a) Beziehung.

Context 1:

Q. camp
A. You may have to get a little... ;

Das Lager ist ein Ort, an dem Menschen leben oder arbeiten. Das Lager kann ein Zeltlager, ein Ferienlager oder ein anderes Lager sein.
Translate the English word to German.:
Therefore, the correct answer is (

0x641:   cmp     esi,ebx                 ;
Therefore, the correct answer is (a) Lager.

Q.
A.

Das ist eine Frage, die nicht so einfach zu beantworten ist. Es gibt

Context 2:

Q. lecture
A. You may have to get a little... ;

Der Vortrag ist eine Rede, die vor einem Publikum gehalten wird. Der Vortrag kann eine Prasentation, ein Vortrag oder eine andere Art von Vortrag sein.
Translate the English word to German.:
Therefore, the correct answer is (
Translate
0x641:   cmp     esi,ebx                 ;
Therefore, the correct answer is (a) Vortrag.

Q.
A.

Der Vortrag ist eine Rede, die vor einem Publikum gehalten wird. Der Vortrag
\end{lstlisting}
\end{tcolorbox}

\subsubsection{Translation English-Spanish}
\begin{tcolorbox}
\begin{lstlisting}[escapechar=@]

Prompt 0:@\textcolor{red}{Traduccion al espanol -> QUOTE: Translate to Spanish}@

Prompt 1: @\textcolor{red}{In Spanish, these five words are:}@

Contexts

Context 0:

Q. bear
A. Translate to Spanish: oso

Q. bear
A. Translate to Spanish: oso

Q. bear
A. Translate to Spanish: oso

Q. bear
A. Translate to Spanish: oso

Q. bear
Translate these five English words to Spanish.
Therefore, the correct answer is (a) oso.

Context 1:
Q. relativity
A. Translate to Spanish: relatividad

Q. relativity
A. Translate to Spanish: relatividad

Q. relativity
A. Translate to Spanish: relatividad

Q. relativity
A. Translate to Spanish: relatividad

Q. relativity
Translate these five English words to Spanish.
Therefore, the correct answer is (a) relatividad.

Q. bear
A. Translate to Spanish: oso

Q. bear
A. Translate to Spanish

Context 2:
Q. concert
A. Translate to Spanish: concierto

Q. concert hall
A. Translate to Spanish: sala de conciertos

Q. concertmaster
A. Translate to Spanish: concertino

Q. concerto
A. Translate to Spanish: concierto

Q.
Translate these five English words to Spanish.
Therefore, the correct answer is (a) concierto.
\end{lstlisting}
\end{tcolorbox}

\subsubsection{Translation English-French }

\begin{lstlisting}[escapechar=@]

Prompt 0: @\textcolor{red}{I've translated 5 words from English to French:}@

Prompt 1: @\textcolor{red}{Translate to French}@

Contexts

Context 0:
Q. destiny
A. I've translated 5 words from English to French:

English
French


destiny
destin

destiny
destin

destiny
destin

destiny
destin

destiny
destin

Translate to French
Therefore, the correct answer is (destin).

Q. destiny
A. I've translated 5 words from English to French:

English

Context 1:
Q. ideology
A. I've translated 5 words from English to French:

English
French

ideology
ideologie

ideology
ideologie

ideology
ideologie

ideology
ideologie

ideology
ideologie


Translate to French
Therefore, the correct answer is (ideologie).

Q. ideology
A. I've translated 5 words from English to French:

English

Context 2:

Q. representation
A. I've translated 5 words from English to French:

English
French

representation
representation

representation
representation

representation
representation


representation
representation


representation
represent
Translate to French
Therefore, the correct answer is (representation).

Q. representation
A. I've translated 5 words from English to French:




English
\end{lstlisting}

\subsubsection{Sentiment Analysis}
\begin{tcolorbox}
\begin{lstlisting}[escapechar=@]

Prompt 0: @\textcolor{red}{Tell if each review was posted by someone with generally positive or negative language}@

Prompt 1:@\textcolor{red}{write the number of positive examples}@

Contexts

Context 0:

Q. Despite its faults, Gangs excels in spectacle and pacing.
A. I gave a friend an instruction and five inputs. The friend read the instruction and wrote an output for every one of the inputs.
B. I gave a friend an instruction and five inputs. The friend read the instruction and wrote an output for every one of the inputs.
C. I gave a friend an instruction and five inputs. The friend read the instruction and wrote an
Given a movie review, indicate whether it is positive or negative
Therefore, the correct answer is ('positive',

Context 1:
Q. I simply can't recommend it enough.
A. I gave a friend an instruction and five inputs. The friend read the instruction and wrote an output for every one of the inputs.
B. I gave a friend an instruction and five inputs. The friend read the instruction and wrote an output for every one of the inputs.
C. I gave a friend an instruction and five inputs. The friend read the instruction and wrote an
write the number of positive examples
Therefore, the correct answer is ('positive',

Context 2:

Q. There's a lot to recommend Read My Lips.
A. I gave a friend an instruction and five inputs. The friend read the instruction and wrote an output for every one of the inputs.
B. I gave a friend an instruction and five inputs. The friend read the instruction and wrote an output for every one of the inputs.
C. I gave a friend an instruction and five inputs. The friend read the instruction and wrote an
write the number of positive examples
Therefore, the correct answer is ('positive',
\end{lstlisting}
\end{tcolorbox}

\subsubsection{Sentence Similarity}
\begin{tcolorbox}
\begin{lstlisting}[escapechar=@]

Prompt 0: @\textcolor{red}{.}@

Prompt 1: !:@\textcolor{red}{There are two kinds of problems / hints that you will see:!<}@

Context 0:

Q. Sentence 1: The polar bear is sliding on the snow. Sentence 2: A polar bear is sliding across the snow.
A. : Read the two sentences and decide how well the second sentence expresses the same idea as the first. Then enter the number of the answer option that best describes your decision.
1. The two sentences express the same idea.
2. The two sentences express the same idea, but the second sentence is not as clear as the first.
3. The two sentences express the same idea, but the second sentence is
: Given two sentences, your algorithm must determine the similarity between the sentences on a scale of 0 to 5. A score of 0 means the sentences have nothing in common. A score of 5 means the sentences are semantically similar. The
Therefore, the correct answer is 5.

Context 1:

Q. Sentence 1: Iranian President praises nuclear deal Sentence 2: Iran arrests several spies near nuclear plant
A. . Output the score according to the following rules:
0: The sentences have nothing in common.
1: The sentences are about the same topic, but express different ideas.
2: The sentences are about the same topic, and express similar ideas.
3: The sentences are about the
 :
Therefore, the correct answer is 1.

Context 2:

Q. Sentence 1: A girl is playing a flute. Sentence 2: A band is playing on a stage.
A. .Output the score according to the following rules:
0: The sentences have nothing in common.
1: The sentences are about the same topic, but express different ideas.
2: The sentences are about the same topic, and express similar ideas
!: There are two kinds of problems / hints that you will see:!<
Therefore, the correct answer is 1.
\end{lstlisting}
\end{tcolorbox}

\subsubsection{Word in Context}
\begin{tcolorbox}
\begin{lstlisting}[escapechar=@]

Prompt 0: @\textcolor{red}{Determine whether the given word is used with the same meaning in both sentences. Write}@

Prompt 1: @\textcolor{red}{Decide whether the given word is used in the same meaning in both sentences.}@

Contexts

Context 0:

Q. Sentence 1: The Times is not the voice of New York. Sentence 2: The voice of the law. Word: voice
A. Determine whether the given word is used with the same meaning in both sentences. Writeyes or no.
The answer is yes.
Decide whether the given word is used in the same meaning in both sentences.
Therefore, the correct answer is (yes).

Context 1:

Q. Sentence 1: Do you communicate well with your advisor? Sentence 2: He and his sons haven't communicated for years. Word: communicate
A. Determine whether the given word is used with the same meaning in both sentences. Writeyes or no.
The answer is yes.
Decide whether the given word is used in the same meaning in both sentences.
Therefore, the correct answer is (yes).

Context 2:

Q. Sentence 1: Can you take me to the main entrance? Sentence 2: Take a scene. Word: take
A. Determine whether the given word is used with the same meaning in both sentences. Writeyes or no.
The answer is no.
Decide whether the given word is used in the same meaning in both sentences.
Therefore, the correct answer is (no).
\end{lstlisting}
\end{tcolorbox}

\section{Ablations}
\label{appendix:ablation}
We performed ablation to measure the impact of various self-referential components of Promptbreeder. We investigated the following mutation operators and mechanisms:
\begin{itemize}
    \item Random initial prompts
    
    The original problem specification for the dataset is used instead of generating an initial task-prompt using the mutation prompt + thinking style + problem specification. 
    \item Random initial mutation prompts
    
    The mutation-prompt "Please summarize and improve the following instruction:" is used instead of randomly selecting a mutation-prompt from the list.
    \item Prompts from context (Lamarckian)
    
    The Lamarckian mutation operator that generates a task-prompt from a correct context is replaced with the default zero-/first-order prompt mutation operation (50:50 chance of one or the other)
    \item Meta-mutation (mutating mutation-prompts)
    
    When meta-mutation would normally take place the default zero-/first-order prompt mutation operation is performed (50:50 chance of one or the other)
\end{itemize}

\begin{figure}[t!]
    \centering
    \includegraphics[width=\textwidth]{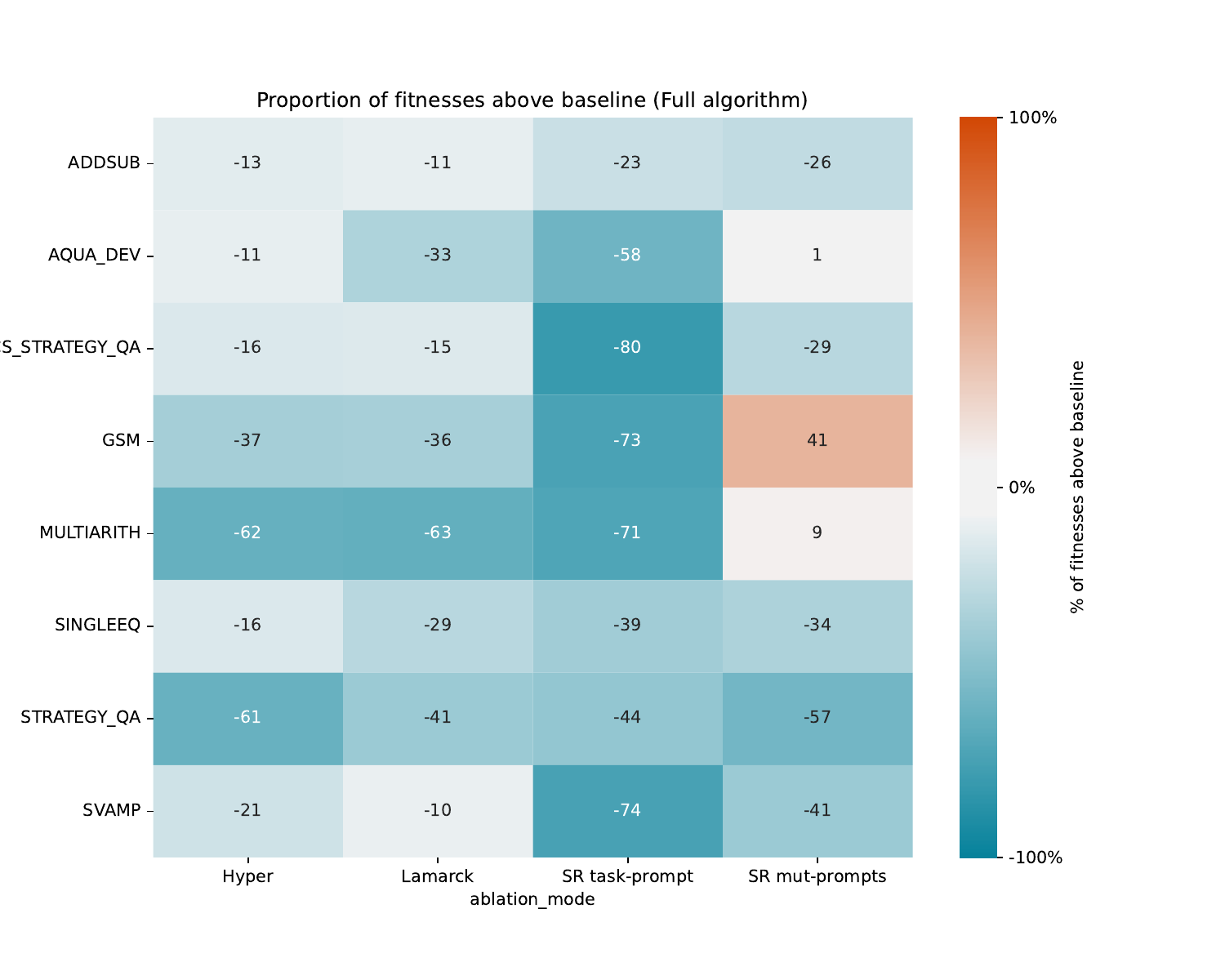}
    \caption{The results of ablating the one by one the self-referential operators compared to using the full algorithm. 0\% signifies an ablated  operation with neither positive nor negative impact. From left to right (Hyper = Removal of mutation-prompt mutation, Lamarck = Removal of Context to task-prompt mutation, SR task-prompt = Removal of thinking-style guided task-prompt initialization, SR mut-prompt = Removal of random selection of a mutation-prompt from the mutation-prompt list.) . Percentage scores close to $-100\%$ indicate that removing the operation results in lower fitness at equivalent points in the run; conversely scores close to $100\%$ mean that the operation is actively harmful, because individuals have higher fitnesses at equivalent points in the run when that operation is removed.}
    \label{fig:ablation_matrix}
\end{figure}

For each dataset and each ablation, we use a population of 10 for 200 evaluations (equivalent to 20 generations, similar to larger experiments in this paper) and compare to the complete algorithm with the same population size and no ablations. To measure how effective an ablated operation is, we determine the proportion of evaluations in the ablation that were higher than the baseline evaluations at each generation, and sum these over all generations in the run. The results in \Cref{fig:ablation_matrix} show that in most cases all the mutation operators have a positive impact on fitness, with the Random Initial Prompts having the largest positive impact across all datasets. 

We also investigated the influence of different mutation operators on the ETHOS hate speech detection dataset~\citep{Mollas_2022} with the under-specified problem specification \prompt{Solve the Problem} (in contrast to the standard problem specification \prompt{Determine whether a text contains hate speech}). Promptbreeder achieved a score of $81.6\%$. The greatest deterioration happens when removing the Lamarckian `from context to prompt' mutation method which induces the instruction from an example of the correct working out ($64.6\%$). The second greatest detriment to performance happens when removing random initialization of mutation prompts, random initialization of prompts, and hyper-mutation of mutation prompts simultaneously, leaving only context mutation ($68.7\%$). Adding back online mutation increases performance back to $70.4\%$ and adding random mutation prompts brings this back up to $73.7\%$. This demonstrates the interplay and importance of Promptbreeder's diverse set of mutation operators. 
\end{document}